\documentclass[10pt,journal,compsoc]{IEEEtran}
\usepackage{amsmath,amsfonts}
\usepackage{algorithmic}
\usepackage{algorithm}
\usepackage{array}
\usepackage[caption=false,font=footnotesize,labelfont=sf,textfont=sf]{subfig}
\usepackage{textcomp}
\usepackage{stfloats}
\usepackage{url}
\usepackage{verbatim}
\usepackage{graphicx}
\usepackage{cite}
\usepackage{pdfpages}
\usepackage{bm}
\usepackage{multirow} 
\usepackage{amsmath}  
\usepackage{amssymb}  
\usepackage{booktabs}
\usepackage{pifont}
\newcommand{\ie}{{\emph{i.e.}}}
\newcommand{\eg}{{\emph{e.g.}}}
\newcommand{\etal}{{\emph{et al.}}}

\newcommand{\xmark}{\ding{55}}

\usepackage{cite}       
\usepackage{hyperref}   

\newcommand{\parhead}[1]{\par\vspace{0.35\baselineskip}\noindent\textbf{#1}}

\begin{document}

\title{Self-Supervised AI-Generated Image Detection: A Camera Metadata Perspective}

\author{Nan~Zhong,
        Mian~Zou,
        Yiran~Xu,
        Zhenxing~Qian,~\IEEEmembership{Senior Member,~IEEE,}
        Xinpeng~Zhang,~\IEEEmembership{Senior Member,~IEEE,}
        Baoyuan~Wu,~\IEEEmembership{Senior Member,~IEEE,}
        and Kede~Ma,~\IEEEmembership{Senior Member,~IEEE}

\thanks{This work was supported in part by the Hong Kong RGC General Research Fund (11220224), the CityU Strategic Research Grants (7005848 and 7005983), and the Guangdong Basic and Applied Basic Research Foundation (2024B1515020095).}
\IEEEcompsocitemizethanks{\IEEEcompsocthanksitem Nan Zhong and Mian Zou are with the Department of Computer Science, City University of Hong Kong, Kowloon, Hong Kong (e-mail: nzhong@cityu.edu.hk, mianzou2-c@my.cityu.edu.hk).
\IEEEcompsocthanksitem Yiran Xu, Zhenxing Qian, and Xinpeng Zhang are with the Department of Computer Science, Fudan University, Shanghai, 200433, China (e-mail: \{yrxu23, zxqian, zhangxinpeng\}@fudan.edu.cn).
\IEEEcompsocthanksitem Baoyuan Wu is with the School of Artificial Intelligence, The Chinese University of Hong Kong, Shenzhen, Guangdong, 518172, China (email: wubaoyuan@cuhk.edu.cn).

\IEEEcompsocthanksitem Kede Ma is with the Department of Computer Science and the Shenzhen Research Institute, City University of Hong Kong, Kowloon, Hong Kong (e-mail: kede.ma@cityu.edu.hk).
}
\thanks{Corresponding Author: Kede Ma.}
}



\IEEEtitleabstractindextext{%
\begin{abstract}
The proliferation of AI-generated imagery poses escalating challenges for multimedia forensics, yet many existing detectors depend on assumptions about the internals of specific generative models, limiting their cross-model applicability. We introduce a self-supervised approach for detecting AI-generated images that leverages camera metadata---specifically
exchangeable image file format (EXIF) tags---to learn features intrinsic to digital photography. Our pretext task trains a feature extractor solely on camera-captured photographs by classifying categorical EXIF tags (\emph{e.g.}, camera model and scene type) and pairwise-ranking ordinal and continuous EXIF tags (\emph{e.g.}, focal length and aperture value). Using these EXIF-induced features, we first perform one-class detection by modeling the distribution of photographic images with a Gaussian mixture model and flagging low-likelihood samples as AI-generated. We then extend to binary detection that treats the learned extractor as a strong regularizer for a classifier of the same architecture, operating on high-frequency residuals from spatially scrambled patches. Extensive experiments across various generative models demonstrate that our EXIF-induced detectors substantially advance the state of the art, delivering strong generalization to in-the-wild samples and robustness to common benign image perturbations. The code and model are publicly available at \url{https://github.com/Ekko-zn/SDAIE}.
\end{abstract}
\begin{IEEEkeywords}
AI-generated image detection, self-supervised learning, image forensics.
\end{IEEEkeywords}}

\maketitle

\section{Introduction}

\IEEEPARstart{AI}{} image generators, spanning autoregressive models~\cite{kingma2016improved,van2016conditional}, variational autoencoders~\cite{vae},  normalizing flows~\cite{kingma2018glow}, generative adversarial networks (GANs)~\cite{goodfellow2014generative, karras2017progressive, brock2018large}, and diffusion models~\cite{dhariwal2021diffusion, gu2022vector,ho2020denoising}, now produce visuals that are widely used but also easily misused. Beyond creative applications, AI-generated imagery can facilitate harassment~\cite{karasavva2021real}, fraud~\cite{de2023unethical}, and large-scale misinformation~\cite{news}, underscoring the need to reliably distinguish camera-captured photographs from AI outputs\footnote{In this work, we refrain from labeling images as ``real'' or ``fake'' when contrasting photographic and AI-generated imagery, as contemporary generative models can memorize and reproduce photographs from their training data~\cite{carlini2023extracting,kadkhodaiegeneralization}. }.

Much of today's progress in AI-generated image detection hinges on model-aware assumptions. For example, methods targeting GANs often exploit upsampling artifacts in spatial~\cite{barni2020cnn} or frequency~\cite{frank2020leveraging} domains, where diffusion-oriented approaches invert or approximate the denoising trajectory and inspect reconstruction errors, similarly in spatial~\cite{wang2023dire} or latent~\cite{cazenavette2024fakeinversion} domains. While effective in-family, such strategies generalize poorly across the rapidly diversifying space of generators and training pipelines.

A more durable route is to avoid learning ``the space of fakes'' altogether~\cite{huh2018fighting}. Ideally, a detector should be trained only on photographs, treating AI-generated images as out-of-distribution anomalies~\cite{chandola2009anomaly,chalapathy2019deep} (also known as one-class classification). Prior related lines of work show promise: projecting into a fixed, semantics-oriented feature space~\cite{ojha2023towards}, checking exchangeable image file format (EXIF) self-consistency across patches~\cite{huh2018fighting}, and aligning photographic images with EXIF viewed as language via contrastive joint embedding~\cite{zheng2023exif}. However, semantic encoders mainly capture high-level associations, while EXIF–language alignment emphasizes cross-patch coherence for splicing; none fully exploit the fine-grained, camera-intrinsic regularities of image formation needed in AI-generated image detection.

In this work, we advocate a self-supervised perspective on EXIF-induced AI-generated image detection. Our key idea is to train a feature extractor from photographic images by predicting camera metadata: multi-class classification for categorical tags (\eg, \texttt{Make}, \texttt{Model}, and \texttt{SceneCaptureType}) and pairwise ranking~\cite{tsai2007frank} for ordinal and continuous tags (\eg, \texttt{FocalLength} and \texttt{ApertureValue}). To emphasize camera-intrinsic characteristics over semantics, the feature extractor operates on high-frequency residuals~\cite{fridrich2012rich} of spatially scrambled patches. This pretext task aims to align the learned features with the physics and camera configurations of digital photography, without relying on any AI-generated examples.

Using the resulting EXIF-induced feature extractor, we instantiate two detectors. The first is a one-class model, SDAIE (
Self-supervised Detection of AI-generated Images using EXIF metadata), which fits a Gaussian mixture model (GMM) to photographic features and flags low-likelihood samples as AI-generated. The second, SDAIE$^\dagger$, is a binary classifier trained with ProGAN~\cite{karras2017progressive} images as a provisional negative class. A representation-alignment regularizer~\cite{li2017learning} ties its intermediate features to the pretext extractor, transferring camera-intrinsic cues and mitigating overfitting to the specified generator. Together, these yield strong performance across diverse generators, including in-the-wild samples from commercial APIs (\eg, Midjourney~\cite{midjourney}), and show robustness under benign post-processing.

In summary, our contributions include
\begin{itemize}
  \item A self-supervised pretext task that leverages EXIF tags to learn camera-intrinsic features from photographs only;
  \item A feature extractor that operates on high-frequency residuals of scrambled patches to suppress semantics and accentuate imaging regularities;
  \item A one-class detector that models photographic features with a GMM and detects anomalies without seeing AI-generated images during training;
  \item A binary detector that uses the pretext extractor as a strong regularizer, improving generalization and robustness across generators and post-processing.
\end{itemize}

\section{Related Work}
In this section, we provide an overview of research topics closely related to ours, including deep generative models, AI-generated image detection, self-supervised learning, and EXIF metadata for computer vision.

\subsection{Deep Generative Models}
Over the past decade, a broad family of deep generative models has emerged,  leveraging a wide range of techniques including autoregression~\cite{kingma2016improved,van2016conditional}, variational inference~\cite{vae}, invertible flow (\ie, Gaussianization)~\cite{kingma2018glow}, adversarial training~\cite{goodfellow2014generative, karras2017progressive, brock2018large}, score matching~\cite{hyvarinen2005estimation}, and stochastic processes~\cite{ho2020denoising}. Among these, GAN-based and diffusion-based models have garnered particular attention for their state-of-the-art image synthesis performance. 

GANs~\cite{goodfellow2014generative} pit a generator against a discriminator in a minmax game. Early
training was notoriously fragile, with
 model collapse, sensitivity to hyperparameters, and unclear progress evaluation. Several advances improved diversity and stability: replacing the Jensen-Shannon divergence with the Wasserstein distance~\cite{arjovsky2017wasserstein}, introducing minibatch discrimination~\cite{salimans2016improved}, and enforcing Lipschitz continuity via spectral normalization~\cite{miyato2018spectral} or gradient penalties~\cite{gulrajani2017improved}. Quantitative metrics such as Fr\'{e}chet inception distance~\cite{heusel2017gans}, inception score~\cite{salimans2016improved}, and distributional precision and recall helped track progress despite limitations. Scaling to high resolutions typically follows a coarse-to-fine recipe, exemplified by progressive growing~\cite{karras2017progressive}, sometimes complemented by super-resolution post-processing. Conditional GANs further enable controllable generation from class labels~\cite{odena2017conditional,nguyen2017plug}, text prompts~\cite{zhang2017stackgan,reed2016generative,hong2018inferring}, or various forms of auxiliary images~\cite{isola2017image}.

More recently, diffusion models~\cite{ho2020denoising} have become the default choice for high-quality, diverse, and stable image synthesis, supported by theoretical insights from score-based modeling~\cite{song2020score}. An image is gradually perturbed to near-white noise in a forward noising process, while a neural network (typically a U-Net~\cite{ronneberger2015u}) is trained to reverse this process step by step. The major practical bottleneck is slow sampling, which has sparked extensive acceleration efforts, including deterministic samplers with input skip connections~\cite{songdenoising}, trajectory distillation~\cite{zheng2023fast,song2023consistency,salimansprogressive}, feature reuse~\cite{ma2024deepcache},  high-order solvers~\cite{karras2022elucidating}, parallel sampling~\cite{shih2024parallel}, and learned sampler optimization~\cite{watson2022learning}. High-resolution synthesis is commonly achieved with latent diffusion~\cite{podell2023sdxl,rombach2022high}, where control signals can be injected via input concatenation, cross-attention, or adaptive normalization. The resulting capabilities have driven the widespread availability of commercial image generation APIs~\cite{dall-e-2, midjourney}, allowing non-experts to produce images from text alone and underscoring the urgent need for reliable detectors of AI-generated imagery.

\subsection{AI-Generated Image Detection}

In step with advances in image generation, early detection techniques focused primarily on GAN-generated content. Durall~\etal~\cite{durall2020watch} showed that up-convolution layers, necessary to many generators, introduce spectrum aliasing that distinguishes synthetic from photographic images. Wang~\etal~\cite{wang2020cnn} further found that a straightforward ResNet-50, trained on photographs and ProGAN~\cite{karras2017progressive} outputs, with simple augmentations (\ie, JPEG compression and Gaussian blurring), transfers well across multiple GAN datasets. Marra \etal~\cite{marra2019gans} observed that architectural choices, training data, and initialization seeds imprint distinctive ``fingerprints'' exploitable for detection and attribution~\cite{yu2019attributing}. However, detectors tailored to GAN artifacts often underperform on high-quality images from diffusion models. To bridge this gap, diffusion-oriented approaches have emerged~\cite{cazenavette2024fakeinversion,luo2024lare,ricker2024aeroblade,wang2023dire}. For instance, DIRE~\cite{wang2023dire} and SeDID~\cite{ma2023exposing} leverage reconstruction errors via diffusion inversion, while DRCT~\cite{chendrct} incorporates reconstructed images as hard negative samples. 
Overall, these lines of work chiefly target artifacts tied to a single model family; as generators evolve, those artifacts drift, diminishing detection reliability. 

A parallel thread seeks some forms of universal fingerprints that generalize across a broad range of generative models, including GANs and diffusion models~\cite{liu2022detecting,tan2023learning,wang2023dire}. Liu~\etal~\cite{liu2022detecting} treated high-frequency noise in photographs, extracted with a pre-trained denoiser, as universal cues; Tan~\etal~\cite{tan2023learning} pursued a related idea using input gradients of a pre-trained classifier. Ojha~\etal~\cite{ojha2023towards} relied solely on CLIP image embeddings with nearest neighbor classification and linear probing. While these strategies improve cross-model generalization, their reliance on semantics-oriented pre-trained encoders limits sensitivity to camera-related cues, particularly for photorealistic diffusion-generated images. We instead introduce a self-supervised approach that learns to classify and rank EXIF metadata associated with photographs, yielding generator-agnostic, EXIF-induced features intrinsic to digital photography for AI-generated image detection.

\subsection{Self-Supervised Learning}
Self-supervised representation learning~\cite{bengio2013representation}, defined as learning without human-provided labels, has become a central paradigm in computer vision. The goal is to train feature extractors that convert raw data into useful representations, thereby simplifying downstream tasks. Arguably, the most important design choice is the pretext task, with prominent families including structural alignment~\cite{doersch2015unsupervised,gidaris2018unsupervised,noroozi2016unsupervised,srivastava2015unsupervised}, clustering~\cite{caron2018deep}, contrastive learning (\eg, instance discrimination~\cite{chen2020simple,he2020momentum} and multi-view learning~\cite{grill2020bootstrap,chen2021exploring}), masked prediction~\cite{pathak2016context,larsson2016learning,he2022masked}, cross-modal alignment~\cite{radford2021learning,arandjelovic2017look}, and generative modeling~\cite{vincent2008extracting}. 

In image forensics, Huh~\etal~\cite{huh2018fighting} detected splicing by testing whether patches within an image share a single imaging pipeline (\ie, exhibit consistent EXIF tags). Zheng~\etal~\cite{zheng2023exif} advanced this idea by learning a joint embedding between image patches and EXIF metadata, treating the latter as a language-like modality. However, standard text tokenizers struggle to encode numerical relationships partly due to token fragmentation, which can limit the effectiveness of EXIF-induced feature learning. In this work, we exploit EXIF information more directly and at finer granularity to train the feature extractor that underpins our detectors, enabling both one-class and binary classification for AI-generated image detection. 

\subsection{EXIF Metadata for Computer Vision}

EXIF is a standardized schema for embedding metadata in image files, providing information about both the imaging device and the capture conditions. Tags such as \texttt{Make}, \texttt{Model}, and \texttt{Software} identify the camera manufacturer, specific model, and any post-processing pipeline. Exposure-related tags (\eg, \texttt{ShutterSpeedValue}, \texttt{ApertureValue}, and \texttt{ISO}) record exposure settings with associated motion characteristics. Geotags (\eg, \texttt{GPSLatitude} and \texttt{GPSLongitude}) localize the capture and supply scene context.
Temporal tags (\eg, \texttt{DateTimeOriginal} and \texttt{SubSecTimeOriginal}) provide precise timestamps, revealing coarse context such as time of day and seasonality. The \texttt{Flash} tag indicates whether and how the flash fired, revealing artificial illumination or low-light conditions. Historically, EXIF has supported camera calibration~\cite{zhang2000flexible} and has been leveraged for image quality assessment~\cite{fang2020perceptual} and image splicing detection~\cite{fan2013estimating,huh2018fighting}. We build on this line of work by exploiting EXIF-induced self-supervised learning for a distinct forensic objective---detecting AI-generated imagery.

\section{Proposed Methods}

\begin{figure*}
    \centering
    \includegraphics[width=.65\linewidth]{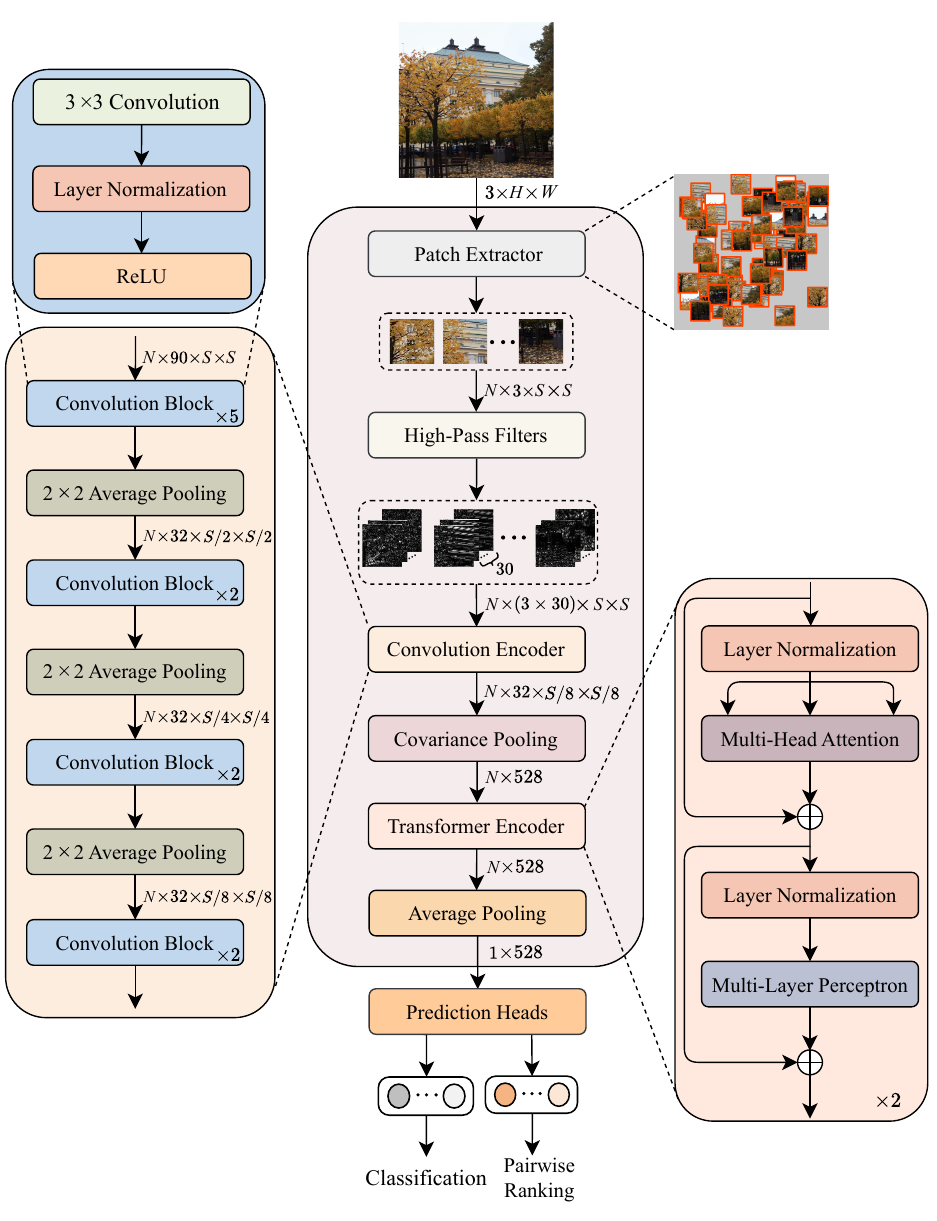}
  \caption{System diagram of the proposed feature extractor using residual patch encoding, covariance pooling, and Transformer attention, trained solely on photographic images with EXIF metadata.}
    \label{fig_framework}
\end{figure*}

We present SDAIE---Self-supervised Detection of AI-generated Images using EXIF metadata. Our central idea is to learn a camera-intrinsic feature extractor (see Fig.~\ref{fig_framework}) only from photographs by predicting their EXIF tags. We then detect AI-generated images as distributional outliers (one-class setting) or via a supervised binary detector regularized by the self-supervised features. This section details the pretext task, the feature extractor, and two detectors built atop the learned representations.

\subsection{Pretext Task}
\begin{table}[]
\caption{Selected EXIF tags for the pretext task.}
\resizebox{\columnwidth}{!}{
\begin{tabular}{lccc}
\toprule
Tag                     & Type        & Example Value                       \\
\midrule
\texttt{Flash}         & Categorical & Flashfired, Flashauto                        \\
\texttt{Make}          & Categorical & Canon, FUJIFILM                                \\
\texttt{MeteringMode}  & Categorical & Multi-segment, Spot                              \\
\texttt{Model}         & Categorical & EOS5DMarkII, EOS7D                            \\
\texttt{SceneCaptureType} & Categorical & Standard, Landscape                        \\
\texttt{ExposureMode}  & Categorical & Auto, Manual                                    \\
\texttt{WhiteBalanceMode} & Categorical & Auto, Manual                                  \\
\texttt{ExposureBiasValue} & Ordinal  & 0 EV, -1 EV                                    \\
\texttt{ISOSpeedRatings}& Ordinal     & 400, 100                                       \\
\texttt{ApertureValue} & Continuous     & F2.8, F4                                       \\
\texttt{ExposureTime}   & Continuous     & 1/60 sec, 1/200 sec                           \\
\texttt{F-Number}      & Continuous     & F2.0, F3.2                                     \\
\texttt{FocalLength}   & Continuous     & 24 mm, 35 mm                                 \\
\texttt{ShutterSpeedValue} & Continuous  & 1/60 sec, 1/100 sec                              \\

\bottomrule
\end{tabular}}
\label{tab_exif_tags}
\end{table}
Modern generators are capable of closely mimicking photographic semantics, so we deliberately avoid semantics and instead align features with imaging regularities recorded in EXIF. From more than one hundred EXIF tags, we retain \textit{fourteen} informative ones with manageable missing rates and high entropy: seven categorical, two ordinal, and five continuous (see Table~\ref{tab_exif_tags}).

For each categorical tag $i$, we attach a lightweight head $\bm g(\cdot;\bm \phi_i)$, parameterized by $\bm \phi_i$, to a shared feature extractor $\bm f(\cdot, 
\bm \theta)$, parameterized by $\bm \theta$, and optimize the cross-entropy loss to predict among the most frequent categories, collapsing the long tail into the ``others'' category:
\begin{align}
\ell_{\mathrm{cat}}(\bm x;\bm \theta,\bm \phi_i) = -\sum_{j=1}^{C_i} p_{ij}(\bm x) \log\left(\sigma_j\left( \bm g \circ \bm f\left(\bm x;\bm\theta,\bm \phi_i\right)\right)\right),
\label{eq_cat}
\end{align}
where
\begin{align}
\sigma_j(\bm z) = \frac{\exp(\bm z_{j})}{\sum_{k=1}^{C_i} \exp(\bm z_k)}.
\label{eq_cat}
\end{align}
Here $C_i$ is the number of categories for the $i$-th EXIF tag, and $p_{ij}(\bm x) \in\{0, 1\}$ represents a one-hot indicator that equals one if the ground-truth category of tag $i$ for photographic image $\bm x$ is $j$. The consolidation of infrequent categories combats heavy class imbalance while preserving discriminative information.

 For each ordinal/continuous tag $i$, we eschew direct regression and instead learn pairwise rankings: given two photos $\bm x$ and $\bm y$ with $i$-th tag values $s_i(\bm x)$ and $s_i(\bm y)$, we predict whether $s_i(\bm x) \ge s_i(\bm y)$:
 \begin{align}
    p_i(\bm x,\bm y) = \begin{cases} 
      1 & s_i(\bm x) \ge s_i(\bm y) \\
      0 & \mathrm{otherwise}.
   \end{cases}
\end{align}
 Using Thurstone’s model~\cite{thurstone1927law}, the predicted probability reduces to a standard Gaussian cumulative distribution function $\Phi(\cdot)$ of the score difference from a scalar prediction head $\bm g(\cdot;\bm \varphi_i)$:
 \begin{align}\label{eq:g0}
{p}(\bm x, \bm y; \bm \theta,\bm \varphi_i) 
&= \Pr\left({\bm g\circ \bm f(\bm x;\bm \theta,\bm \varphi_i) \ge \bm g\circ \bm f(\bm y; \bm \theta,\bm \varphi_i)}\right) \nonumber \\
&= \Phi\left(\frac{\bm g\circ \bm f(\bm x;\bm \theta,\bm \varphi_i)-\bm g\circ \bm f(\bm y;\bm \theta,\bm \varphi_i)}{\sqrt{2}}\right),
\end{align}
 where we fix the variance parameter to one (corresponding to Thurstone's Case V model). As with categorical tags, we adopt the cross-entropy loss for optimization:
\begin{align}
\ell_{\mathrm{rank}}(\bm{x}, \bm{y}; \bm{\theta}, \bm \varphi_i) =& - p_i(\bm{x}, \bm{y}) \log({p}(\bm{x}, \bm{y};\bm \theta,\bm \varphi_i))  \nonumber \\
 &- (1 - p_i(\bm{x}, \bm{y})) \log(1 - {p}(\bm{x}, \bm{y};\bm \theta; \bm \varphi_i)).
\label{eq_ord}
\end{align}
Notably, this ranking formulation is fairly robust to uneven spacings and mis-specified quantization, typical of EXIF numerics. The total pretext objective linearly combines tag-wise classification and pairwise ranking losses:
\begin{align}
    \ell(\mathcal{B};\bm \theta,\bm \phi,\bm \varphi) = & \frac{1}{\vert\mathcal{B}\vert}\sum_{\bm x \in \mathcal{B}}\sum_{i}\alpha_i\ell_{\mathrm{cat}}(\bm x;\bm \theta,\bm \phi_i)  \nonumber\\
    &+\frac{1}{\binom{\vert\mathcal{B}\vert}{2}}\sum_{(\bm x, \bm y) \in \mathcal{B}}\sum_{i}\beta_i\ell_{\mathrm{rank}}(\bm x,\bm y;\bm \theta,\bm \varphi_i), 
\end{align}
where $\mathcal{B}$ is a  minibatch of photographic images, $i$ indexes EXIF tags, and $\{\bm \alpha, \bm \beta\}$ are predetermined weighting vectors.

\begin{figure*}[]
    \centering
    \includegraphics[page=15,trim=45mm 150mm 60mm 0mm,clip, width=0.9 \textwidth]{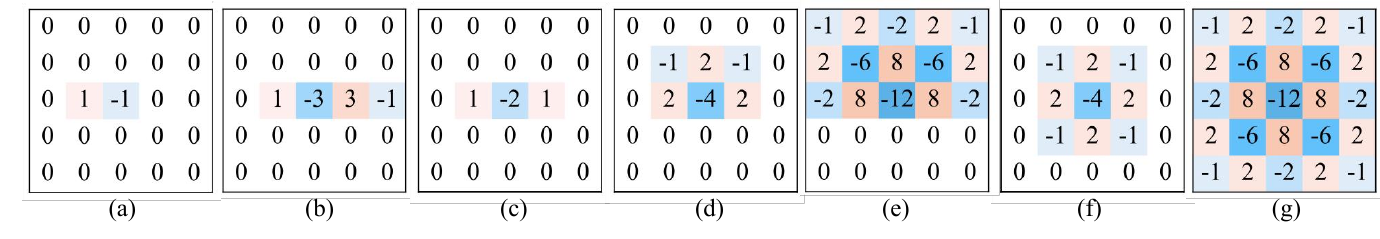}
    \caption{Seven prototype kernels for constructing the high-pass filter bank via discrete rotations.  (a) and (b) are rotated to eight compass directions $\{\nearrow, \rightarrow, \searrow, \downarrow, \swarrow, \leftarrow, \nwarrow, \uparrow\}$; (c) is rotated to four directions $\{\rightarrow, \downarrow, \nearrow, \searrow\}$ (opposite directions are equivalent); (d) and (e) are rotated to the four cardinal directions $\{\rightarrow, \downarrow, \leftarrow, \uparrow\}$; and (f) and (g) are used without rotation. In total, this yields $30$ high-pass filters ($2\times8 + 1\times4 + 2\times4 + 2$).}
    \label{fig_hpf}
\end{figure*}

\subsection{Network Architecture Design}
AI-generated image detection---a specialized task in image forensics---relies less on image semantics and more on camera-intrinsic, low-level cues that directly inform image synthesis. To emphasize such cues, given a photographic image $\bm x \in \mathbb{R}^{3\times H\times W}$, we first extract multiple (and possibly overlapping) patches $\mathbf X\in\mathbb{R}^{N\times 3\times S\times S}$, where $(H, W)$ and $(S, S)$ denote the image and patch sizes, and $N$ is the number of extracted patches. We intentionally discard patch coordinates (\ie, no positional embeddings~\cite{SU2024127063}), producing a patch-scrambling effect that disrupts scene structure and reduces the utility of high-level context.
Consistent with  perceptual and psychophysical findings~\cite{stojanoski2014time,foulsham2011scrambled}, such scrambling shifts reliance toward low-level features like color and texture to make sense of scene structure. To further emphasize forensic microstructures rather than semantics, we apply the high-pass filters of Fridrich and Kodovsk\'{y}~\cite{fridrich2012rich} to each patch (see Fig.~\ref{fig_hpf}), amplifying residual signals that carry traces of the in-camera pipeline, including sensor noise and photo-response non-uniformity, color filter array demosaicing periodicity, lens harpening, and compression footprints. The filtered patches are then passed through several stages of convolution, layer normalization, ReLU activation, and $2\times 2$ average pooling to extract local features. 

After convolution encoding, we summarize patch features with covariance pooling~\cite{li2018towards} rather than average pooling. This is because average pooling primarily captures first-order statistics (\ie, means), which are weak or near-zero in high-pass residuals and thus poorly informative. In contrast, covariance pooling preserves second-order structure that is more stable to global luminance/contrast shifts and semantic changes than means. Moreover, these second-order cues better capture the ``texture of residuals'' that separates digital acquisition from synthesis. Finally, we apply a Transformer encoder with self-attention across patches to model long-range interactions, followed by average pooling, yielding a $528$-dimensional feature representation.

\begin{figure}[t]
    \centering
    \captionsetup[subfloat]{
    labelformat=empty
    } 
    \subfloat[\footnotesize SDXL]{
        \includegraphics[page=4,trim=110mm 135mm 110mm 0mm,clip,width=0.45\textwidth]{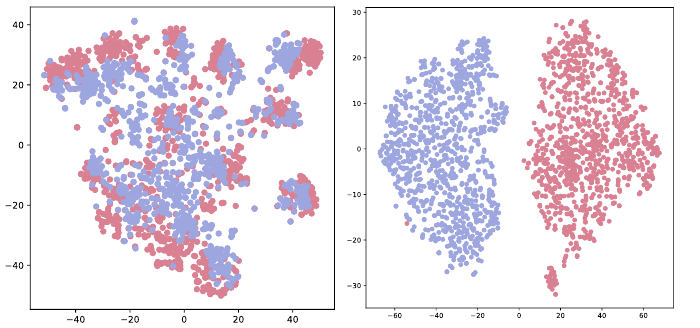}
    }
    \\
    \vspace{-10pt}
    \subfloat[\footnotesize Midjourney ]{
        \includegraphics[page=5,trim=110mm 135mm 110mm 0mm,clip,width=0.45\textwidth]{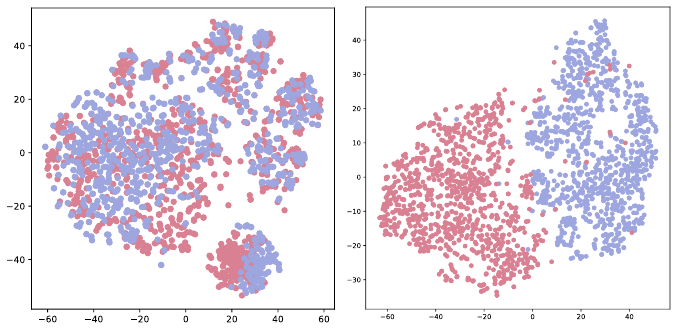}
    }
    \\
    \vspace{-10pt}
    \subfloat[\footnotesize DALLE2]{
        \includegraphics[page=6,trim=110mm 135mm 110mm 0mm,clip,width=0.45\textwidth]{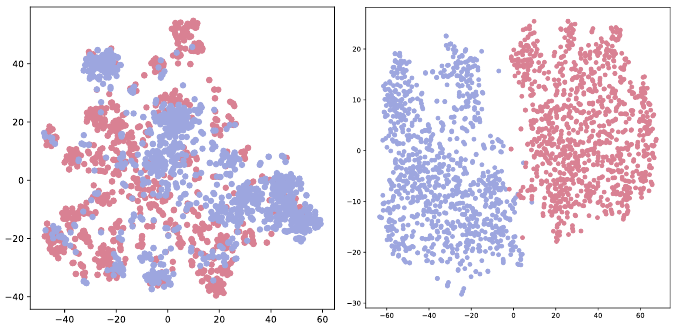}
    }
    \caption{t-SNE visualization~\cite{van2008visualizing} of feature spaces: CLIP~\cite{radford2021learning} (left) versus our EXIF-induced extractor (right), contrasting photographic (red) and AI-generated (blue) images.}
    \label{fig_tsne_motivation}
\end{figure}

 During training, we append fourteen lightweight prediction heads, each implemented by a fully connected layer corresponding to classification or ranking objectives for individual EXIF tags. These heads are used only for pretext supervision and are discarded at test time for AI-generated image detection. Fig.~\ref{fig_tsne_motivation} compares t-SNE embeddings~\cite{van2008visualizing} of features from CLIP~\cite{radford2021learning} and from our EXIF-induced extractor for photographic images versus those generated by 
 SDXL~\cite{podell2023sdxl}, Midjourney~\cite{midjourney}, and DALLE2 \cite{dall-e-2}. Due to their photorealism and semantic richness, AI-generated images cluster near photographic images in the CLIP space. In contrast, our EXIF-induced features produce a clear separation between the two domains, a notable result given that the feature extractor is trained exclusively on photographs.

\begin{figure*}
    \centering
    \subfloat[SDAIE]{\includegraphics[width=0.5\linewidth]{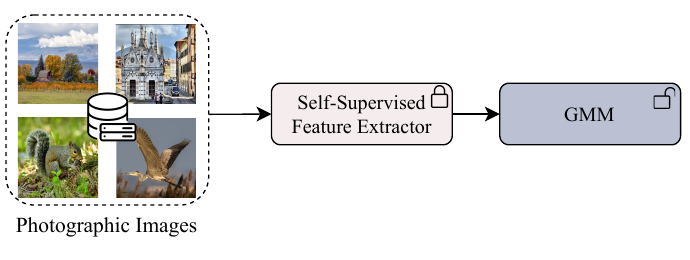}}
    \subfloat[SDAIE$^\dagger$]{\includegraphics[width=0.495\linewidth]{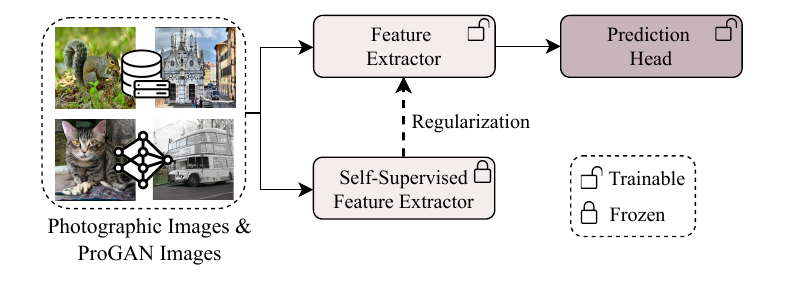}}
    \caption{Overview of SDAIE and SDAIE$^\dagger$. 
    \textbf{(a)} One-class detection: modeling EXIF-induced photographic features with a GMM. \textbf{(b)} Binary detection: regularizing the classifier by the pretext feature extractor to preserve camera-intrinsic cues.}
    \label{fig: sdaie_sdaie_dagger}
\end{figure*}

\subsection{SDAIE}
In this subsection, we address the problem of flagging AI-generated images as anomalies when only photographic images are available for training (see Fig.~\ref{fig: sdaie_sdaie_dagger}(a)). Let $\bm v$ denote the EXIF-induced feature vector of a photographic image. We model the feature distribution with a GMM~\cite{mclachlan1988mixture}, which is expressive (with mixtures capturing multi-modal device/setting/shooting variability), interpretable (with component means/covariances), and probabilistic (enabling principled uncertainty and outlier scoring):
\begin{align}
p(\bm v; \bm \omega) = \sum_{k = 1}^{K} \pi_{k} \mathcal{N}\left(\bm v ; \bm \mu_{k}, \bm \Sigma_{k}\right),
\end{align}
where $\bm \omega =\{\pi_{k},\bm \mu_k,  \bm \Sigma_k\}_{k=1}^K$, $\pi_k \ge 0$ and $\sum_k \pi_k =1$. Parameters are estimated by maximizing the training log-likelihood via expectation-maximization~\cite{moon1996expectation}. Given training features $\{\bm v^{(i)}\}_{i=1}^M$, the E-step computes responsibilities:
\begin{align}
q^{(i)}_{k} = \frac{\pi_k \mathcal{N}(\bm v^{(i)} ; \bm \mu_k, \bm \Sigma_k)}{\sum_{k' = 1}^{K} \pi_{k'} \mathcal{N}(\bm v^{(i)} ; \bm \mu_{k'}, \bm \Sigma_{k'})}. 
\end{align}
With $M_k =\sum_{i=1}^M q^{(i)}_k$, the M-step updates are
\begin{align}
\pi_k = \frac{M_k}{M}, 
\end{align}
\begin{align}
\bm \mu_k = \frac{1}{M_k}\sum_{i=1}^{M} q^{(i)}_{k} \bm v^{(i)}, 
\end{align}
\begin{align}
\bm \Sigma_k = \frac{1}{M_k}\sum_{i=1}^{M} q^{(i)}_{k} (\bm v^{(i)} - \bm \mu_k)(\bm v^{(i)} - \bm \mu_k)^\intercal.
\end{align}
At inference, an image is scored by its log-likelihood $s(\bm v) = \log p (\bm v; \bm \omega)$. Photographic images similar to the training modes receive high scores, whereas AI-generated images tend to have low scores under the learned distribution. We set a decision threshold $\tau$ to the $\rho$-th quantile of training log-likelihoods and predict ``AI-generated'' whenever $s(\bm v) < \tau$. Generally, we choose a small $\tau$ to control the false alarm rate on photographic images.

\subsection{SDAIE$^\dagger$}

To further improve detection accuracy, we revisit the widely used binary classification paradigm for AI-generated image detection~\cite{wang2020cnn,frank2020leveraging,wang2023dire}. In this setting, images synthesized by one or more generative models are treated as negative samples, while photographic images serve as positive samples (aligning with SDAIE). However, as discussed in the Introduction, the breadth and rapid evolution of generative models make it unlikely that a supervised detector trained on a finite set of generators can achieve sufficient coverage and robust generalization to unseen models.

 To mitigate overfitting to the training generators, we draw on the idea of learning without forgetting~\cite{Decontinuallearning} and regularize the classifier using the self-supervised feature extractor. Specifically, we cache intermediate and final feature representations of training images computed by the self-supervised extractor, denoted $\{\{\bm v_l(\bm x^{(i)};\bm \theta^\star)\}_{l\in\mathcal{L}}\}_{i=1}^M$, where $\bm \theta^\star$ are the extractor parameters learned via the EXIF-induced self-supervised task, and $\mathcal{L}$ indexes the stored feature stages. With the original heads removed, the network is equipped with a new prediction head $\bm g(\cdot;\bm \psi)$, comprising a fully connected layer followed by a sigmoid. Training jointly optimizes  $\bm \theta$ and $\bm \psi$ to  maximize detection accuracy while retaining EXIF-aligned features. The detection loss is the standard cross-entropy:
\begin{align}
\ell_{\mathrm{cls}}(\bm x;\bm \theta,\bm \psi) = & - p(\bm x) \log\left( \bm g(\bm v(\bm x;\bm \theta);\bm \psi) \right) \nonumber\\
& -  (1-p(\bm x)) \log\left( 1-  \bm g(\bm v(\bm x;\bm \theta);\bm \psi)\right),
\end{align}
where $p(\bm x)=1$ for photographic images and  $p(\bm x) = 0$ otherwise. Feature preservation is enforced by an $\ell_2$-induced matching regularizer that transfers camera-intrinsic knowledge from the pretext feature extractor:

\begin{align}
\ell_{\mathrm{reg}}(\bm x;\bm \theta) = \frac{1}{\vert\mathcal{L}\vert}\sum_{l\in\mathcal{L}} \frac{1}{D_l}\left\Vert \bm v_l(\bm x;\bm \theta^\star) - \bm v_l(\bm x;\bm \theta)\right\Vert_2^2,
\end{align}
where $D_l=\mathrm{dim}(\bm v_l)$ returns the dimension of the (flattened) representation at layer $l$. Per-layer normalization by 
$D_l$ and average over $l$
make the penalty scale-invariant across layers. Finally, the minibatch loss combines the two terms:
\begin{align}
\ell(\mathcal{B};\bm \theta,\bm \psi) = \frac{1}{\vert\mathcal{B}\vert}\sum_{\bm x\in \mathcal{B}}\bigg(\ell_{\mathrm{cls}}(\bm x;\bm \theta,\bm \psi) + \gamma  \ell_{\mathrm{reg}}(\bm x;\bm \theta)\bigg),
\label{eq_binary_classification}
\end{align}
where $\gamma$ controls the trade-off between detection accuracy and feature consistency. 

Importantly, EXIF metadata is used only in the self-supervised stage, encouraging EXIF-aligned features beneficial for detecting AI-generated images. In both SDAIE and SDAIE$^\dagger$ training, as well as at inference, EXIF fields are neither accessed nor required. In other words, predictions rely solely on image pixels. Consequently, any removal, corruption, or manipulation of EXIF metadata at test time has no effect on model performance.

\section{Experiments}
In this section, we conduct comprehensive experiments to assess both one-class SDAIE and binary SDAIE$^\dagger$ detectors. We first describe the experimental setups, including datasets, competing methods, and hyperparameter configurations. We then evaluate detection performance across a wide range of generative models and conclude with ablation studies, isolating the contributions of key components.

\subsection{Experimental Setups}\label{subsec:es}

\begin{table}[t]

\caption{Details of evaluation benchmarks.}
  \centering
  \scalebox{1.0}{
    \begin{tabular}{lccc}
    \toprule
    \multirow{2}{*}{Generator} & Image & Image & Photographic \\
    &        Size    &     Number       &     Source \\

    \midrule
    ProGAN \cite{karras2017progressive} & $256\times256$ & 8.0k  & LSUN \\
    StyleGAN \cite{karras2019style} & $256\times256$ & 12.0k  & LSUN \\
    BigGAN \cite{brock2018large} & $256\times256$ & 4.0k  & ImageNet \\
    CycleGAN \cite{zhu2017unpaired} & $256\times256$ & 2.6k  & ImageNet \\
    StarGAN \cite{choi2018stargan} & $256\times256$ & 4.0k  & CelebA \\
    GauGAN \cite{park2019semantic} & $256\times256$ & 10.0k & COCO \\
    StyleGAN2 \cite{karras2020analyzing} & $256\times256$ & 15.9k & LSUN \\
    WFIR \cite{whichfaceisreal} & 1024×1024 & 2.0k  & FFHQ \\
    ADM \cite{dhariwal2021diffusion}  & $256\times256$ & 12.0k & ImageNet \\
    Glide \cite{nichol2021glide} & $256\times256$ & 12.0k & ImageNet \\
    Midjourney \cite{midjourney} & $1,024\times1,024$ & 12.0k & ImageNet \\
    SDv1.4 \cite{rombach2022high} & $512\times512$ & 12.0k & ImageNet \\
    SDv1.5 \cite{rombach2022high} & $512\times512$ & 16.0k & ImageNet \\
    VQDM \cite{gu2022vector} & $256\times256$ & 12.0k & ImageNet \\
    WUKONG \cite{wukong} & $512\times512$ & 12.0k & ImageNet \\
    DALLE2 \cite{dall-e-2} & $256\times256$ & 2.0k    & ImageNet \\
    SDXL \cite{podell2023sdxl}  & $1,024\times1,024$ & 4.0k    & COCO \\
    \bottomrule
    \end{tabular}}
  \label{tab_dataset}
\end{table}

\begin{figure*}
    \hspace*{-4em}  
    \begin{minipage}{\dimexpr\textwidth+8em}  
    \centering
    \captionsetup[subfloat]{labelformat=empty}

    \subfloat[\footnotesize ProGAN]{\includegraphics[width=0.12\textwidth]{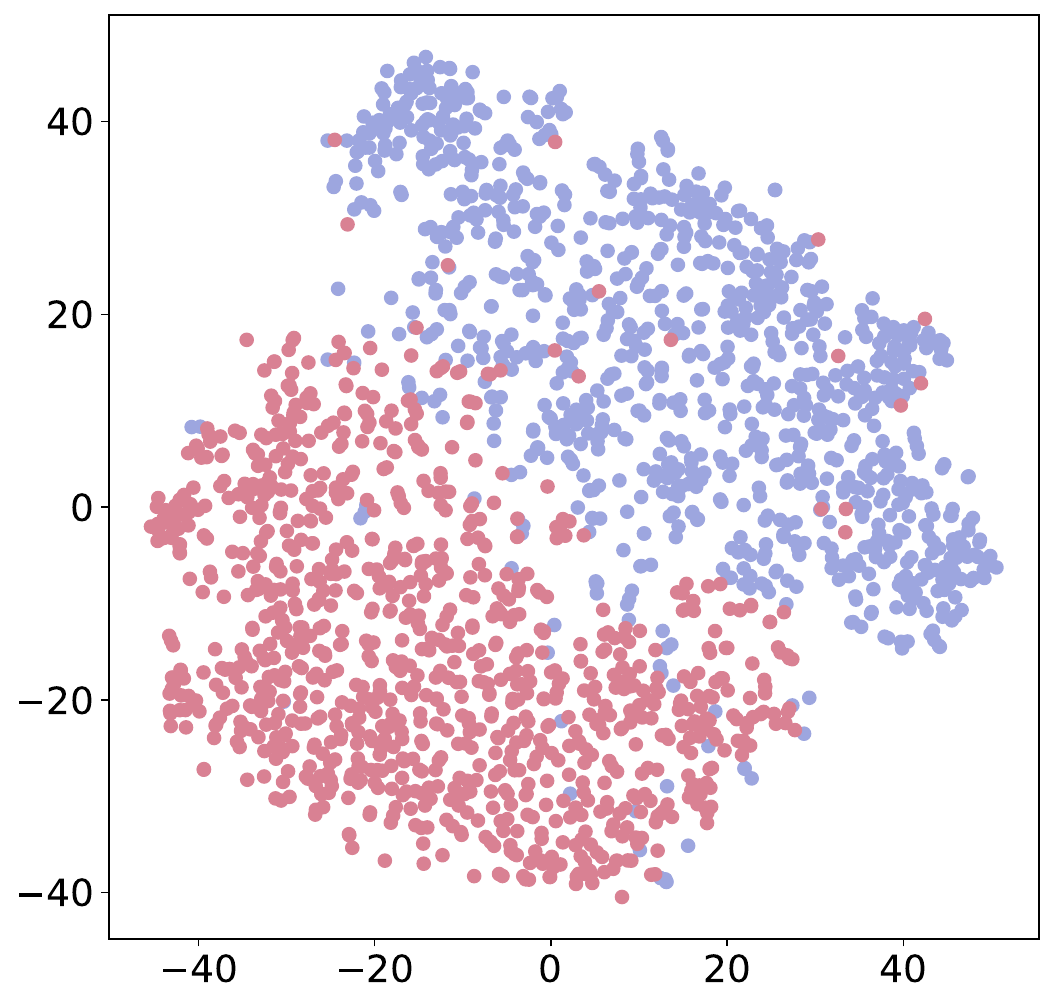}} \hspace{0.015\textwidth}
    \subfloat[\footnotesize StyleGAN]{\includegraphics[width=0.12\textwidth]{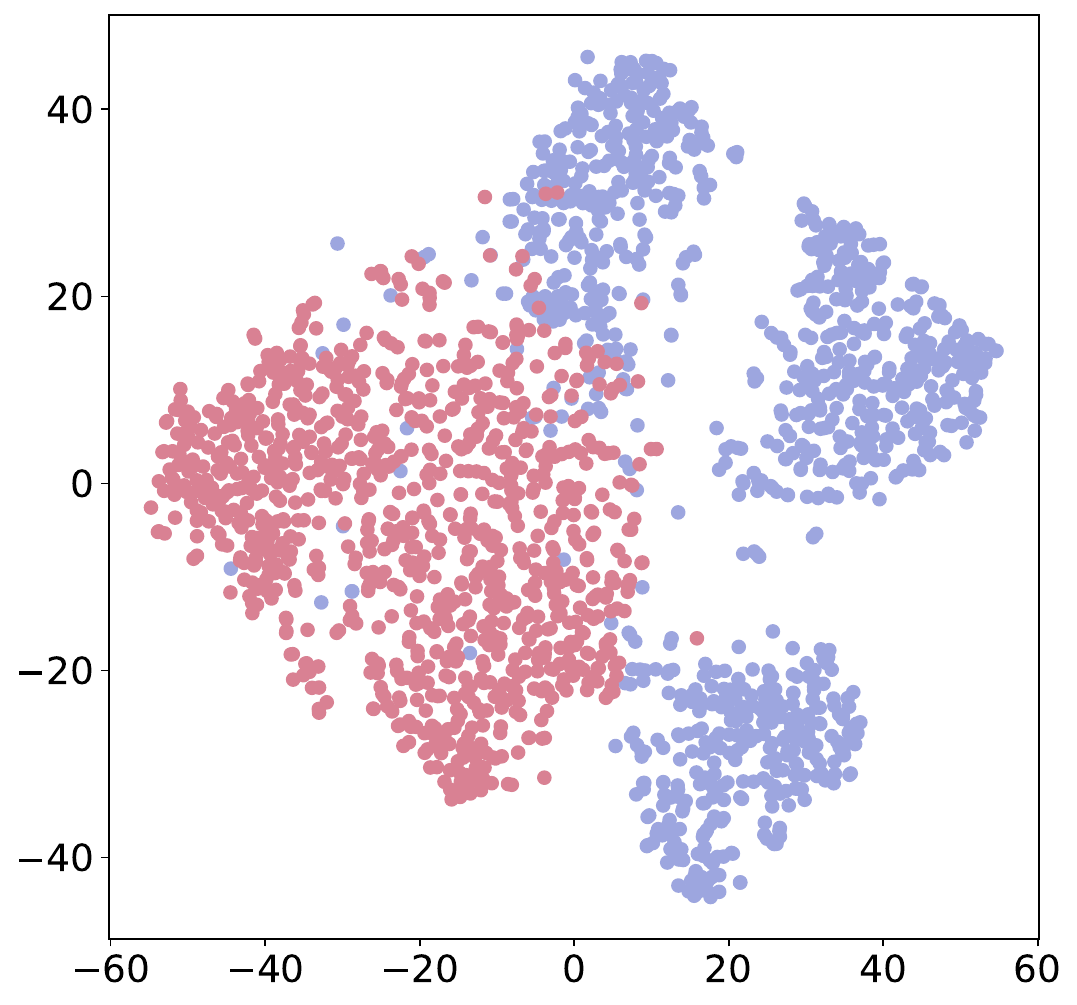}} \hspace{0.015\textwidth}
    \subfloat[\footnotesize BigGAN]{\includegraphics[width=0.12\textwidth]{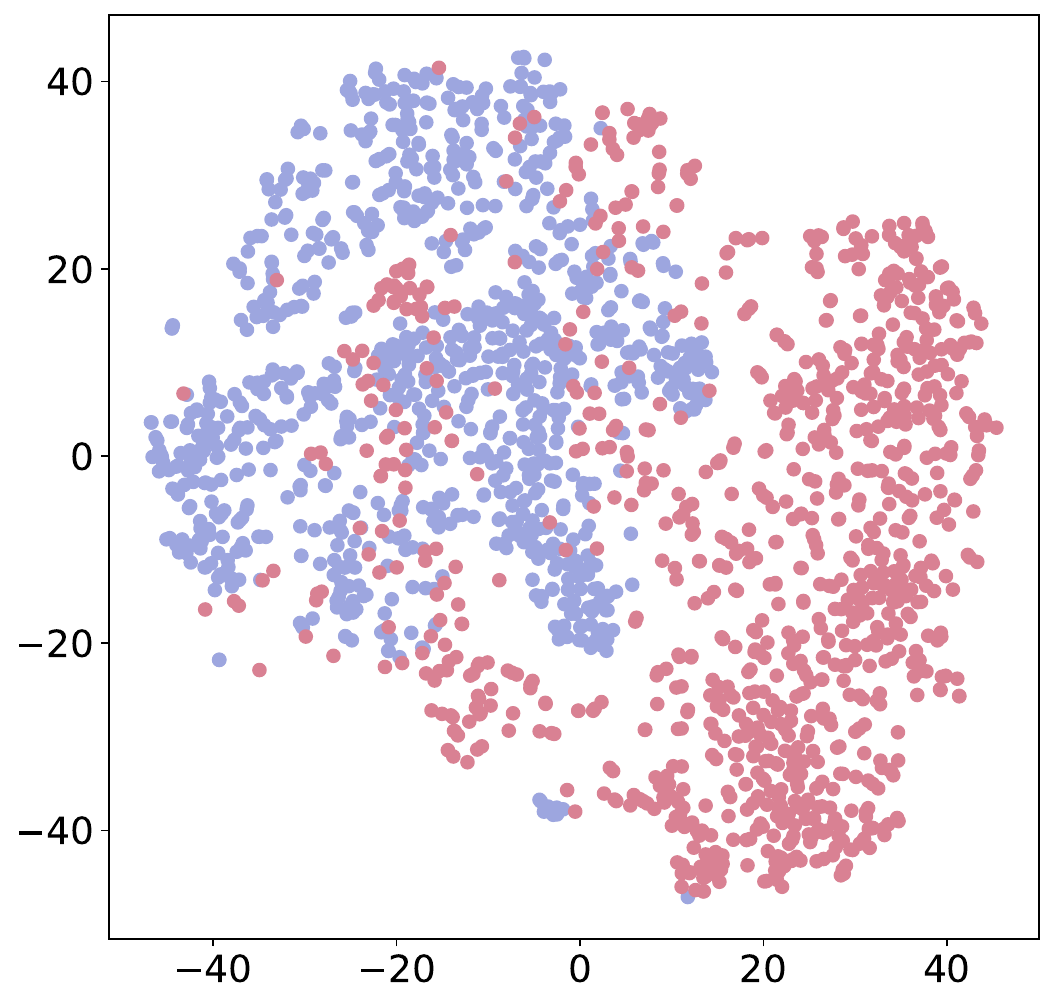}} \hspace{0.015\textwidth}
    \subfloat[\footnotesize CycleGAN]{\includegraphics[width=0.12\textwidth]{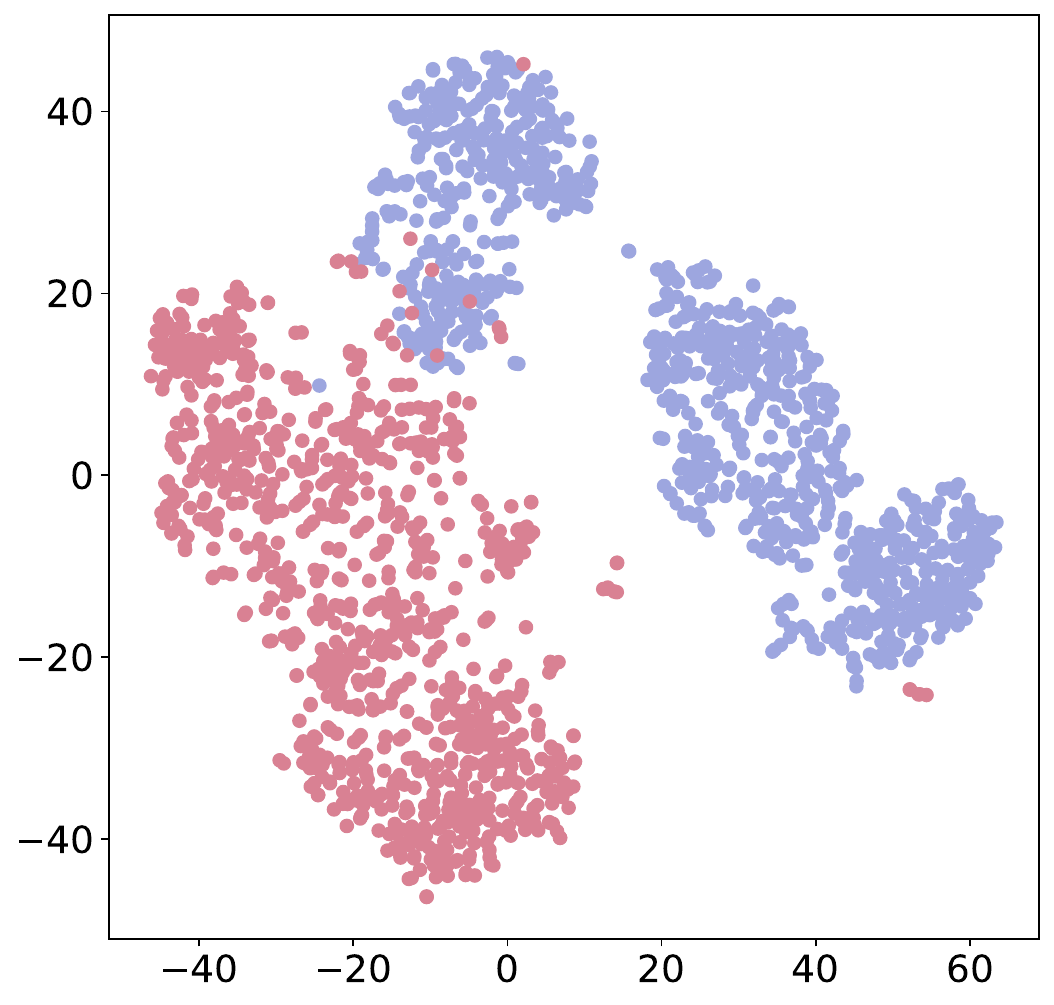}} \hspace{0.015\textwidth}
    \subfloat[\footnotesize StarGAN]{\includegraphics[width=0.12\textwidth]{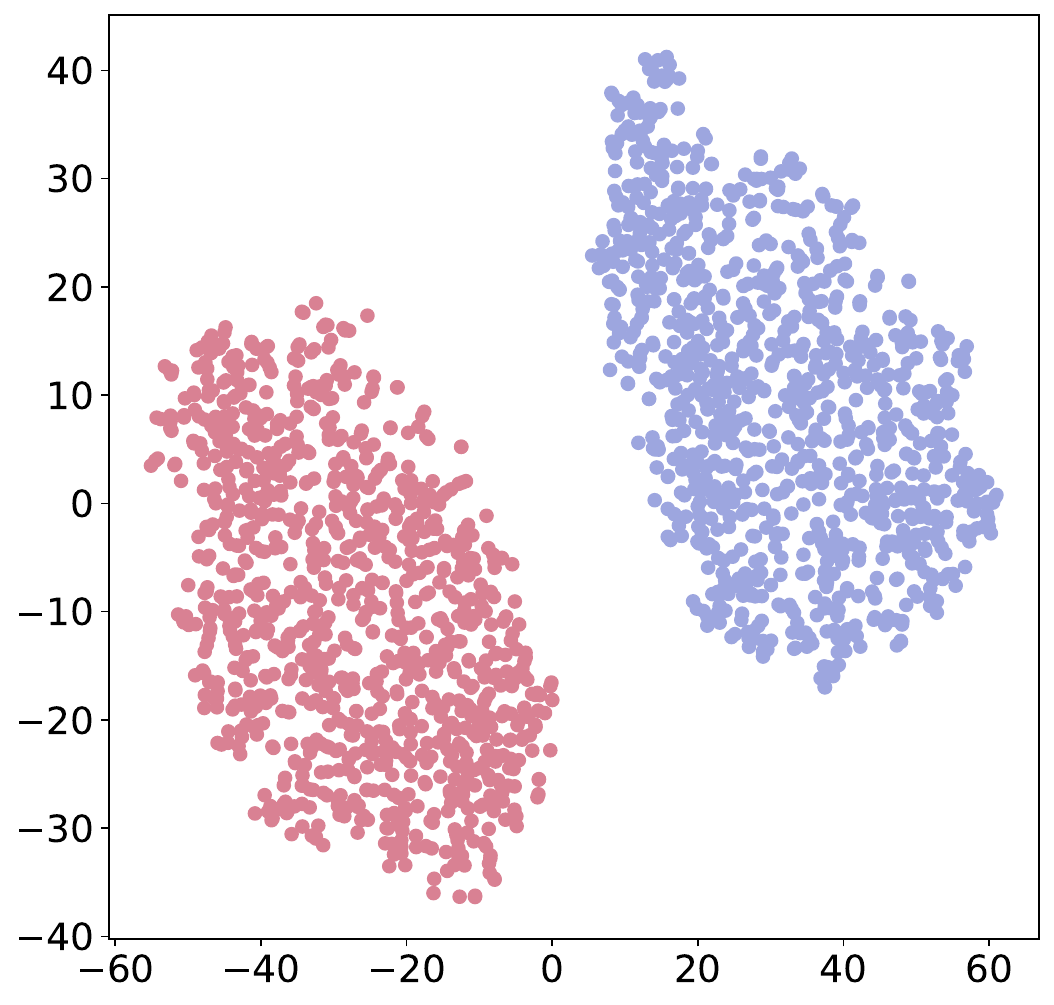}} \hspace{0.015\textwidth}
    \subfloat[\footnotesize GauGAN]{\includegraphics[width=0.12\textwidth]{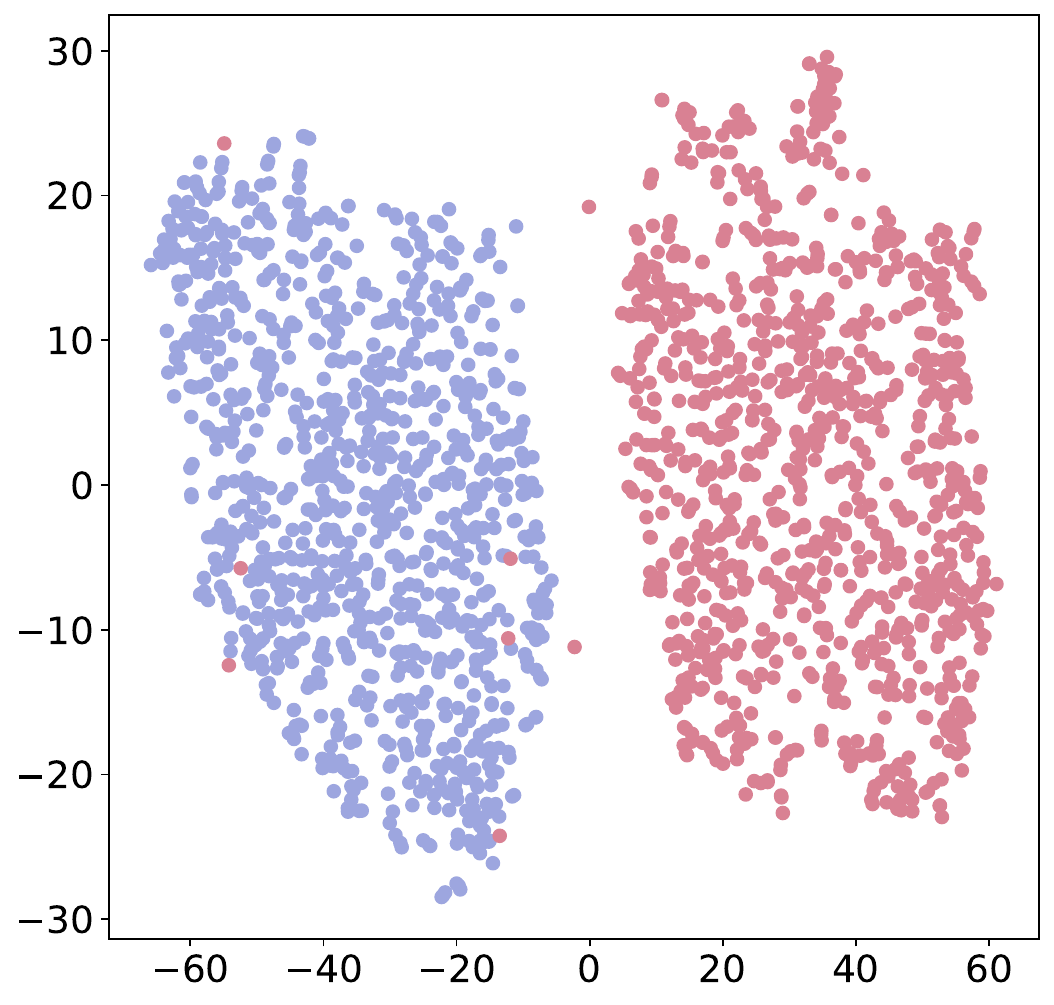}}
    \\[1pt]

    \subfloat[\footnotesize StyleGAN2]{\includegraphics[width=0.12\textwidth]{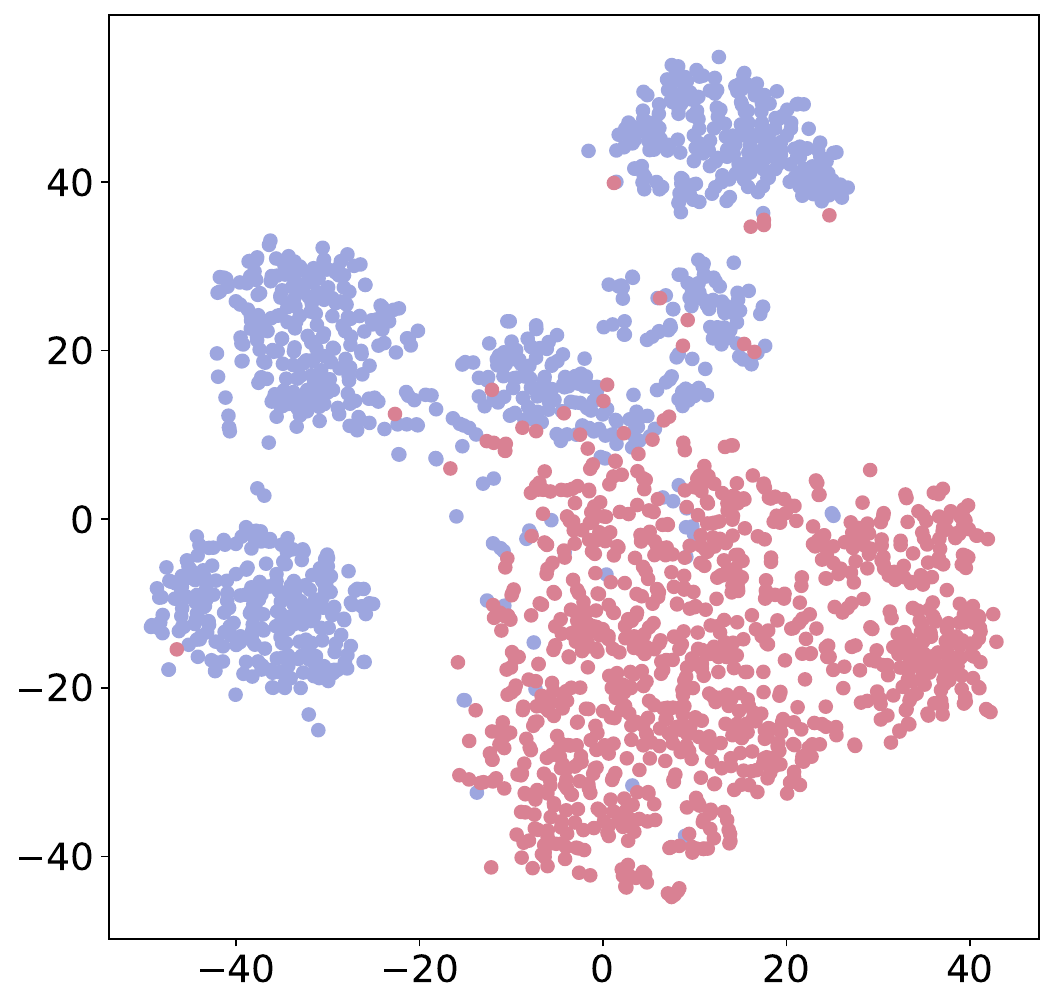}} \hspace{0.015\textwidth}
    \subfloat[\footnotesize WFIR]{\includegraphics[width=0.12\textwidth]{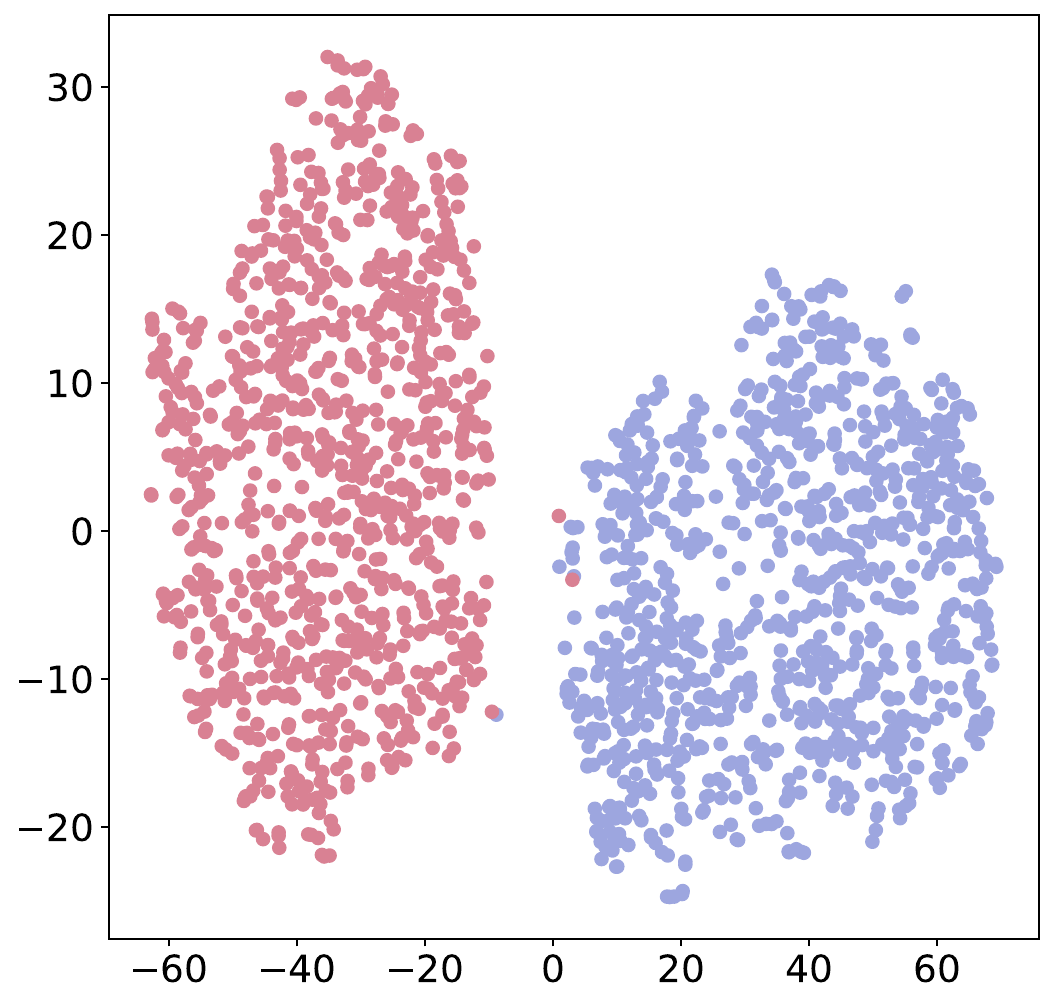}} \hspace{0.015\textwidth}
    \subfloat[\footnotesize ADM]{\includegraphics[width=0.12\textwidth]{likelihood/ADM.pdf}} \hspace{0.015\textwidth}
    \subfloat[\footnotesize Glide]{\includegraphics[width=0.12\textwidth]{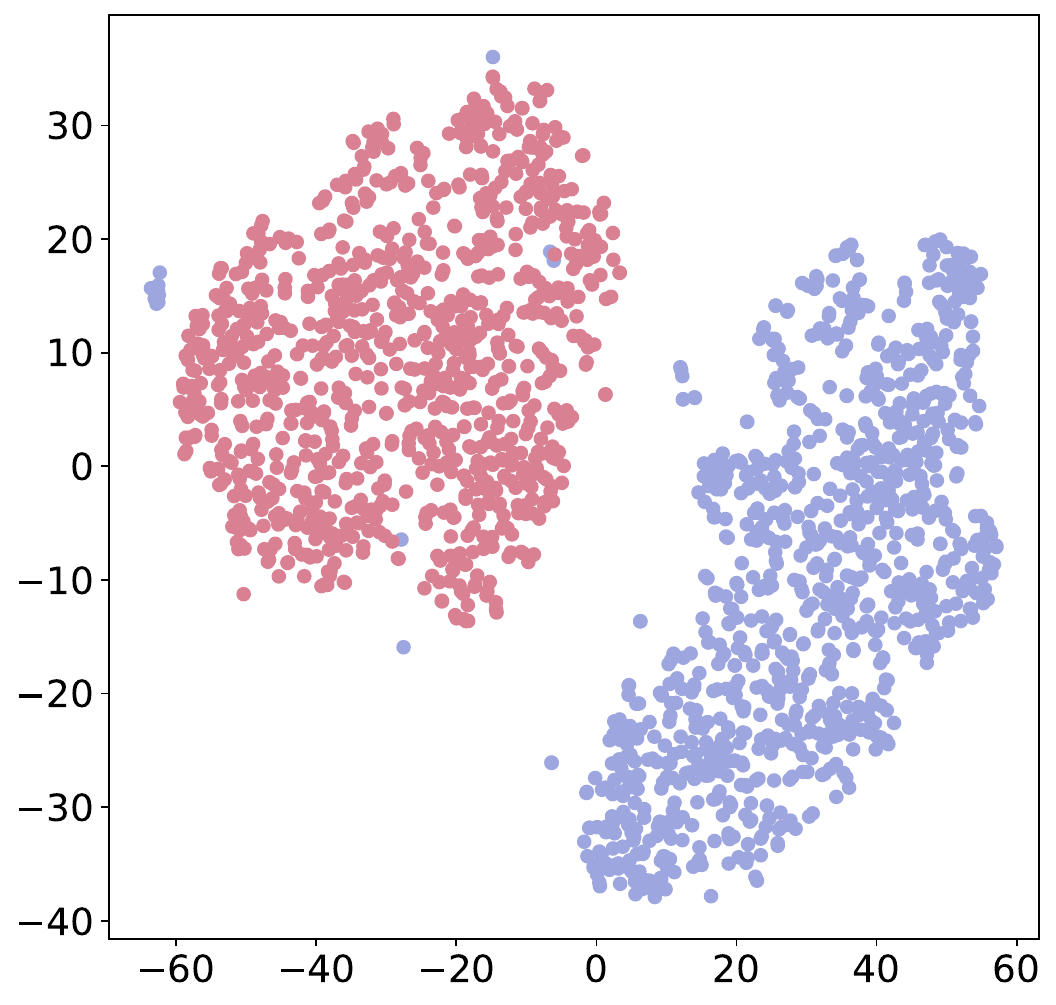}} \hspace{0.015\textwidth}
    \subfloat[\footnotesize Midjourney]{\includegraphics[width=0.12\textwidth]{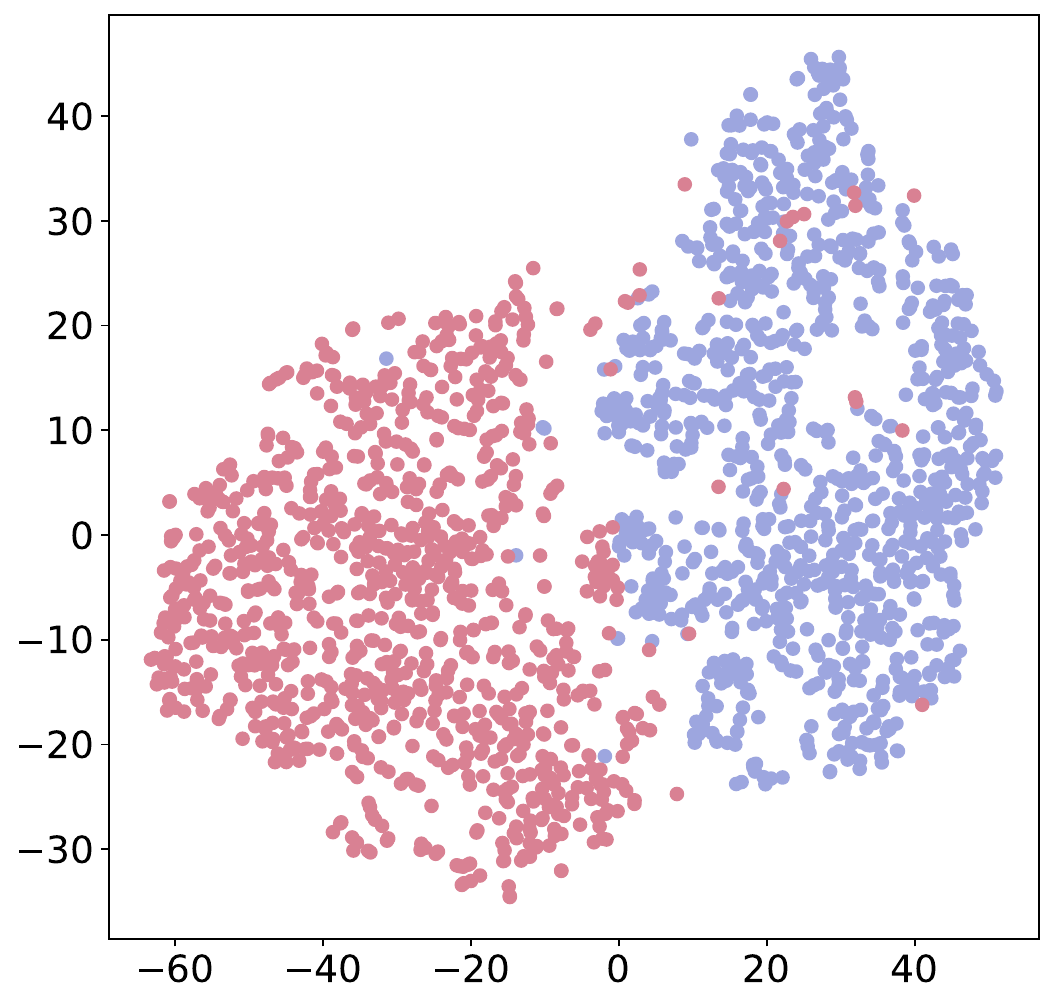}} \hspace{0.015\textwidth}
    \subfloat[\footnotesize SDv1.4]{\includegraphics[width=0.12\textwidth]{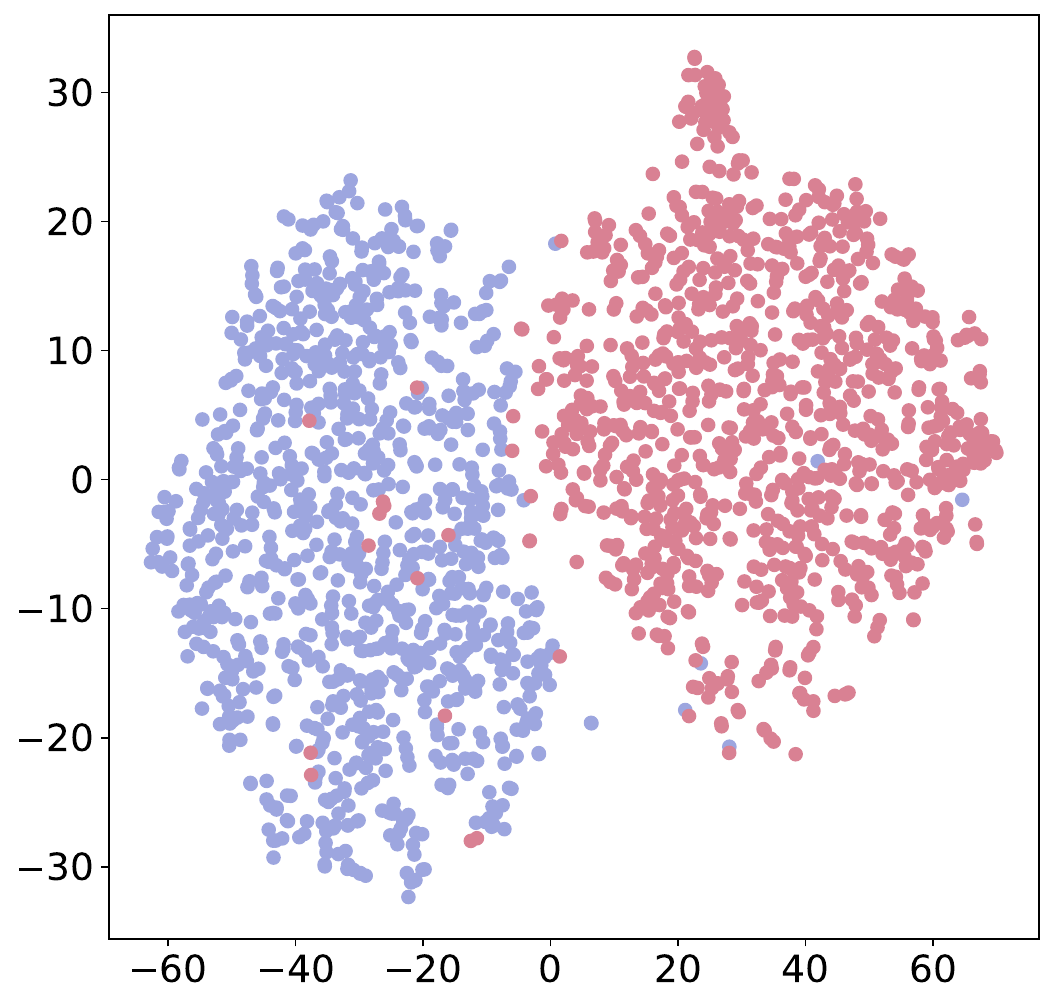}}
    \\[1pt]

    \hspace*{\fill}
    \subfloat[\footnotesize SDv1.5]{\includegraphics[width=0.12\textwidth]{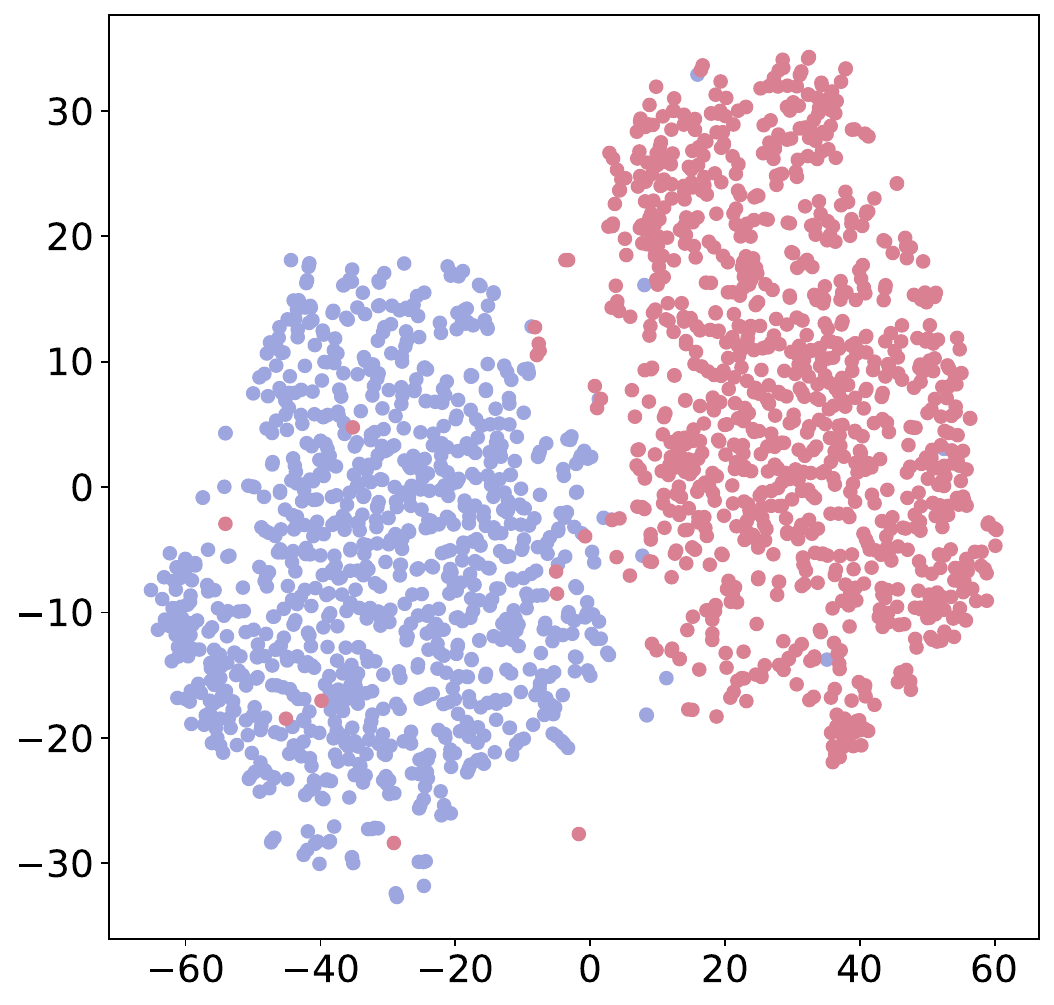}} \hspace{0.005\textwidth}
    \subfloat[\footnotesize VQDM]{\includegraphics[width=0.12\textwidth]{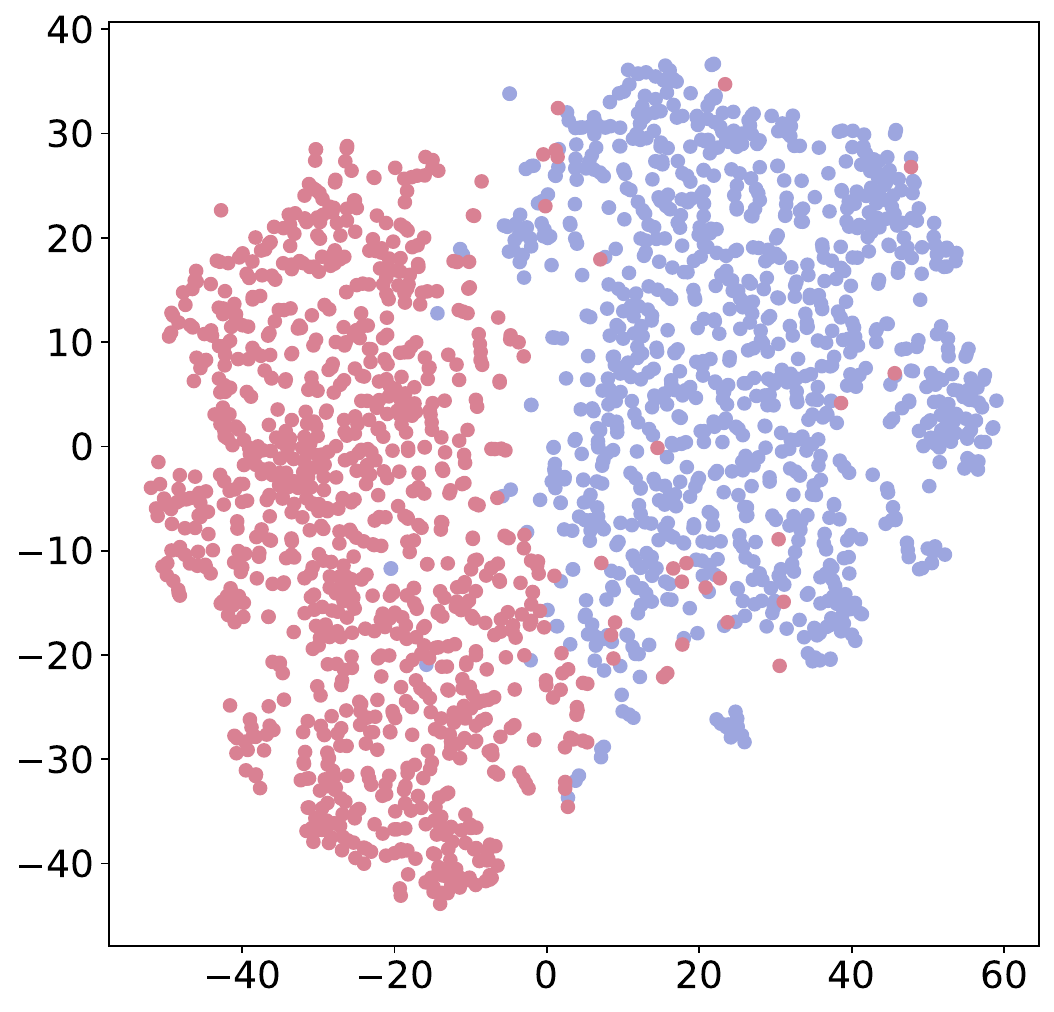}} \hspace{0.005\textwidth}
    \subfloat[\footnotesize WUKONG]{\includegraphics[width=0.12\textwidth]{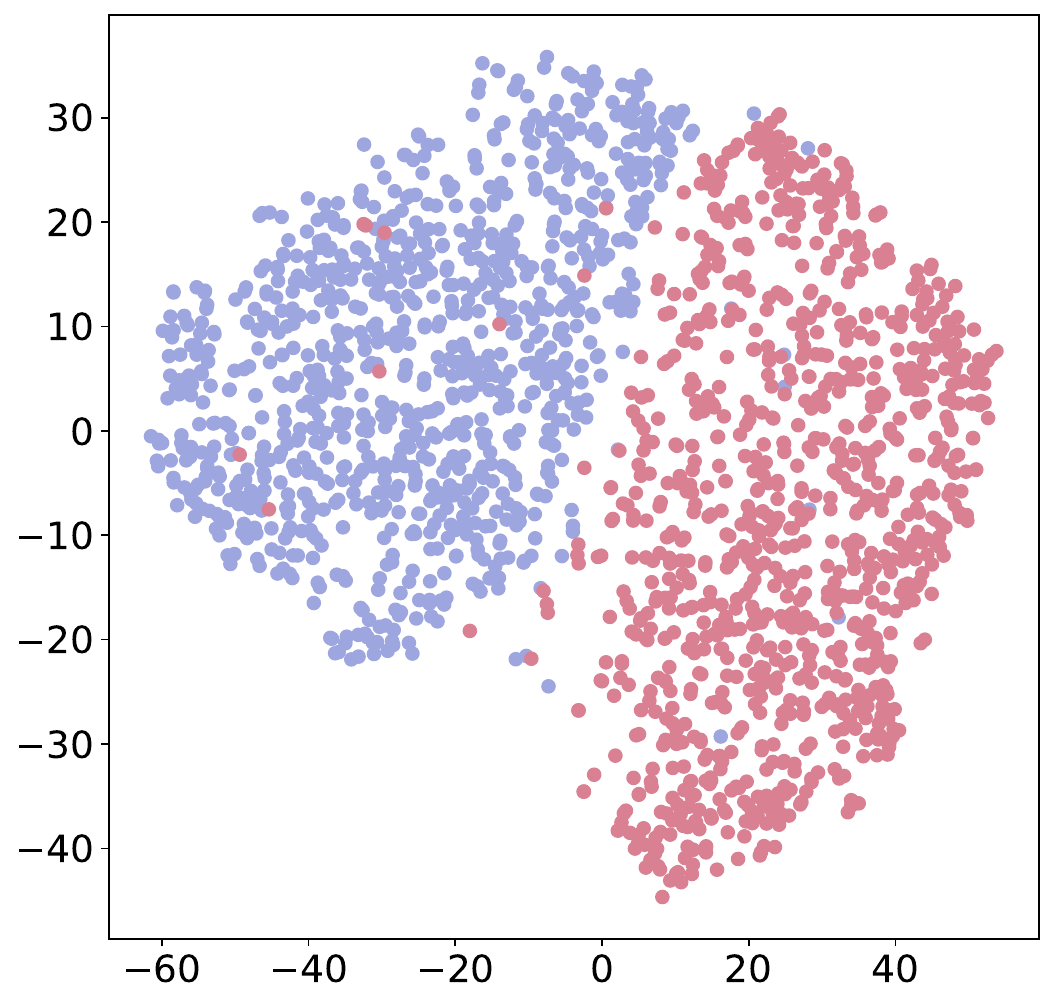}} \hspace{0.005\textwidth}
    \subfloat[\footnotesize DALLE2]{\includegraphics[width=0.12\textwidth]{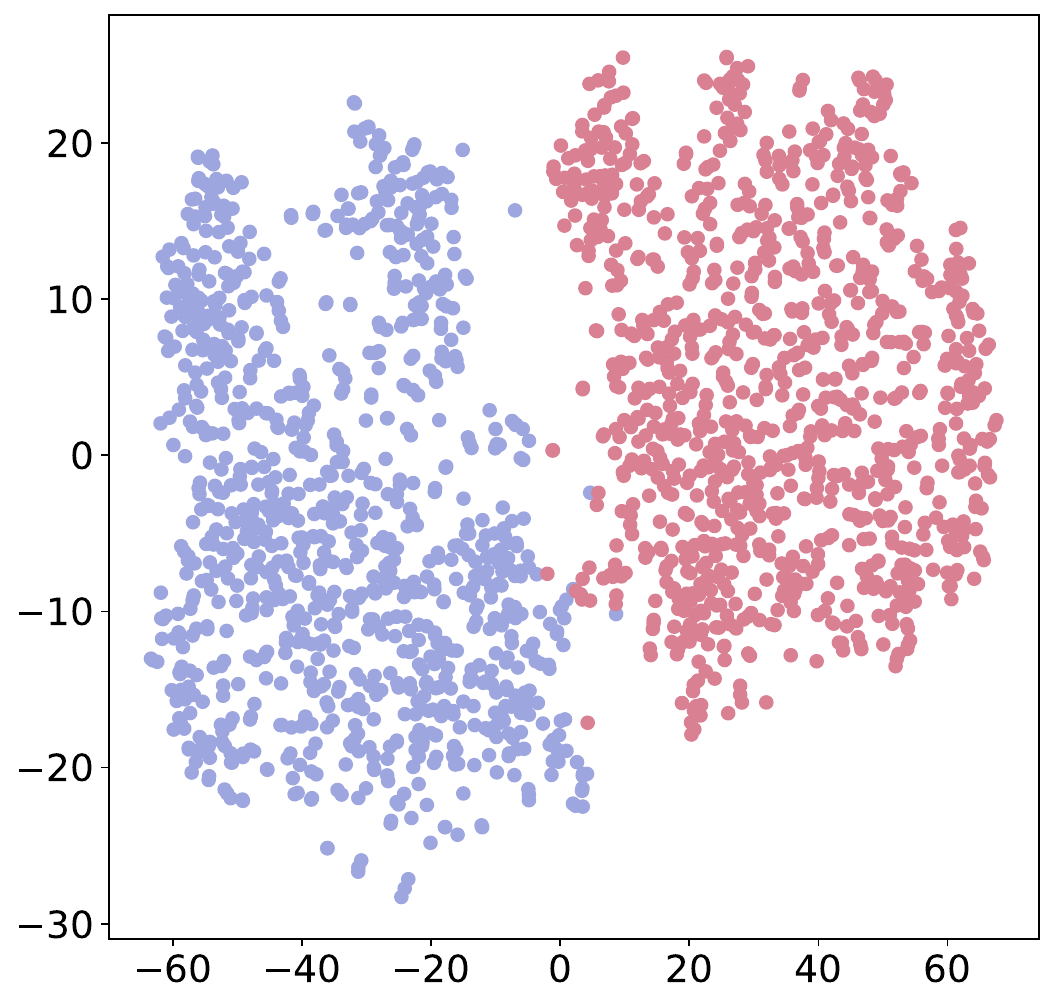}} \hspace{0.005\textwidth}
    \subfloat[\footnotesize SDXL]{\includegraphics[width=0.12\textwidth]{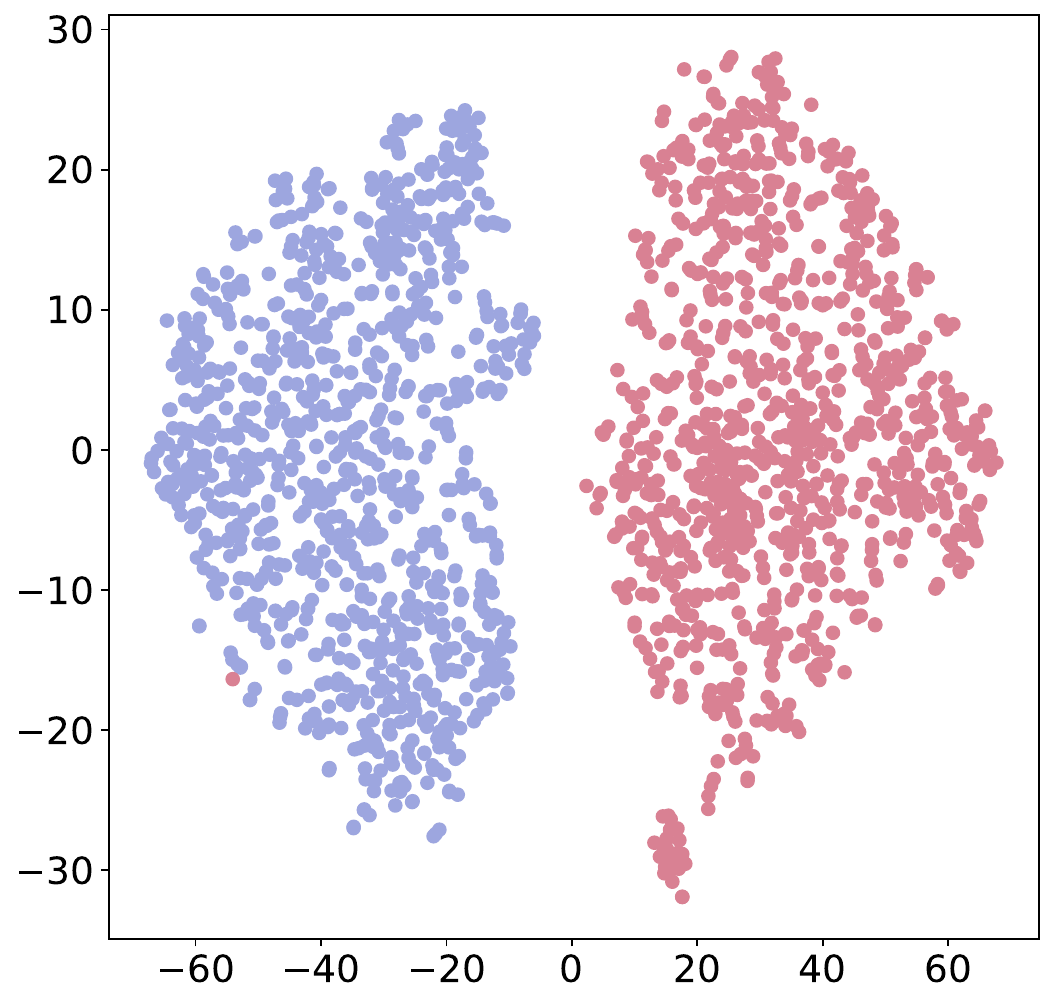}}
    \hspace*{\fill}
    
    \end{minipage}

    \caption{t-SNE visualization of EXIF-induced features showing a clear separation between photographic (red) and AI-generated (blue) images.}
    \label{fig_tsne}
\end{figure*}

\parhead{Datasets.} We construct our training and evaluation data as follows. For  pre-training the EXIF-induced feature extractor, we randomly sample $0.93$M photographic images from YFCC100M~\cite{yfcc}, retaining only samples whose EXIF metadata contains the fourteen tags listed in Table~\ref{tab_exif_tags}. For one-class SDAIE, the GMM is fitted  on a mixed set of $10$k photographic images drawn from ImageNet~\cite{deng2009imagenet} and LSUN~\cite{yu2015lsun}. For binary SDAIE$^\dagger$, we follow established practice~\cite{wang2020cnn,tan2023learning,ojha2023towards} and use a $720$k training set, comprising $360$k LSUN photographs and $360$k ProGAN-generated images~\cite{karras2017progressive}. For main experiments, we assess generalization across
seventeen generators:
1) ProGAN~\cite{karras2017progressive}, 2) StyleGAN~\cite{karras2019style}, 3) BigGAN~\cite{brock2018large}, 4) CycleGAN~\cite{zhu2017unpaired}, 5) StarGAN~\cite{choi2018stargan}, 6) GauGAN~\cite{park2019semantic}, 7) StyleGAN2~\cite{karras2020analyzing}, 8) WFIR~\cite{whichfaceisreal}, 9) ADM~\cite{dhariwal2021diffusion}, 10) Glide~\cite{nichol2021glide}, 11) Midjourney~\cite{midjourney}, 12) SDv1.4~\cite{rombach2022high}, 13) SDv1.5~\cite{rombach2022high}, 14) VQDM~\cite{gu2022vector}, 15) WUKONG~\cite{wukong}, 16) DALLE2~\cite{dall-e-2}, and 17) SDXL~\cite{podell2023sdxl} (more details in Table \ref{tab_dataset}).

\parhead{Competing Methods.} AI-generated image detection spans diverse modeling assumptions and feature biases. To ensure broad and fair coverage, we compare against nine representative detectors selected for methodological diversity (\eg, texture/spectrum cues, local–global fusion, denoiser-based residuals, discriminator-gradient features, diffusion-oriented reconstruction signals, and semantics-driven embeddings), strong reported performance, and complementary sensitivity to GAN- and diffusion-era artifacts.
\begin{enumerate}
    \item CNNSpot~\cite{wang2020cnn} (CVPR'20) adopts a plain ResNet-50 with simple augmentations (\eg, JPEG compression) to improve generalization. Owing to its simplicity and effectiveness, it is widely used as a baseline.
    \item GramNet~\cite{liu2020global} (CVPR'20) strengthens texture modeling by inserting Gram operators into a ResNet-like architecture. 
    \item Frank20~\cite{frank2020leveraging} (ICML'20) distinguishes photographic from AI-generated images by contrasting their frequency spectra.
    \item Ju22~\cite{ju2022fusing} (ICIP'22) integrates local and global representations to enhance detection.
    \item LNP~\cite{liu2022detecting} (ECCV'22) exposes AI-generated imagery via noise residuals extracted from a pre-trained denoiser.
    \item LGrad~\cite{tan2023learning} (CVPR'23) uses gradients of the StyleGAN discriminator with respect to the input image as features, which are classified by a ResNet50.
    \item DIRE~\cite{wang2023dire} (ICCV'23) targets diffusion-generated images by measuring reconstruction errors along forward and reverse diffusion trajectories.
    \item UnivFD~\cite{ojha2023towards} (CVPR'23) builds on a pre-trained CLIP image encoder to leverage strong semantic representations for visual comprehension and detection.
    \item NPR~\cite{tan2024rethinking} (CVPR'24) detects structural artifacts induced by ubiquitous upsampling operators through analysis of neighboring pixel statistics.
\end{enumerate}

\begin{figure*}
    \hspace*{-4em}  
    \begin{minipage}{\dimexpr\textwidth+8em}  
    \centering
    \captionsetup[subfloat]{labelformat=empty}

    \subfloat[\footnotesize ProGAN]{\includegraphics[width=0.12\textwidth]{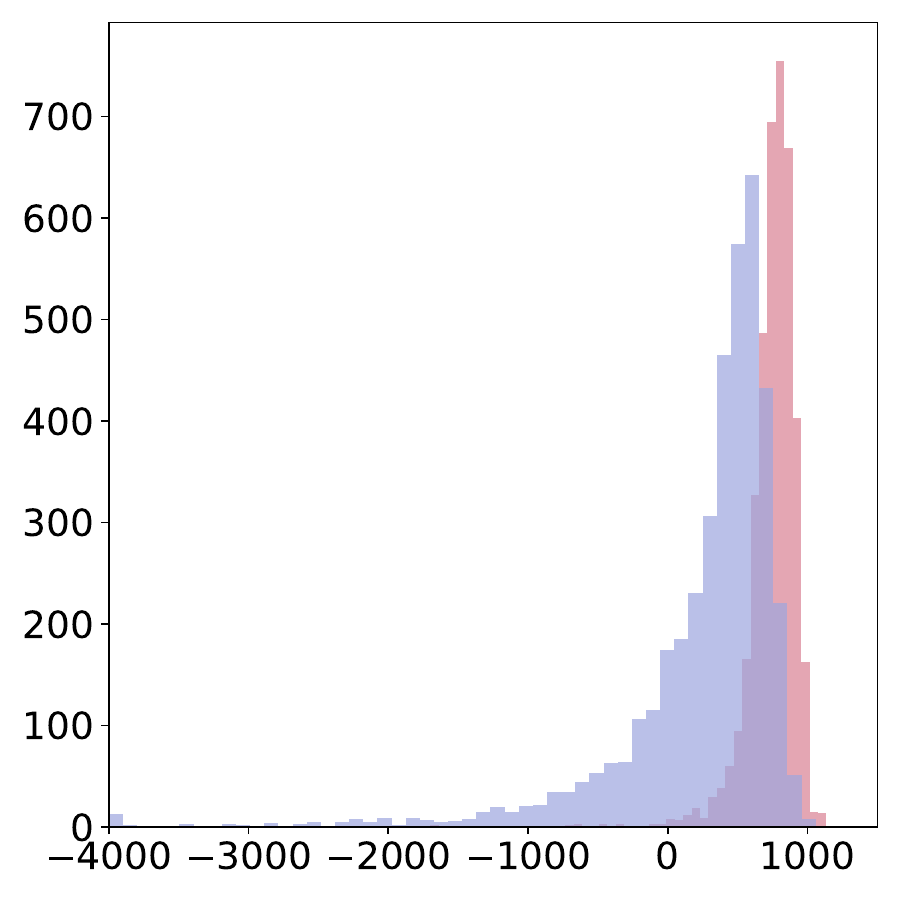}} \hspace{0.015\textwidth}
    \subfloat[\footnotesize StyleGAN]{\includegraphics[width=0.12\textwidth]{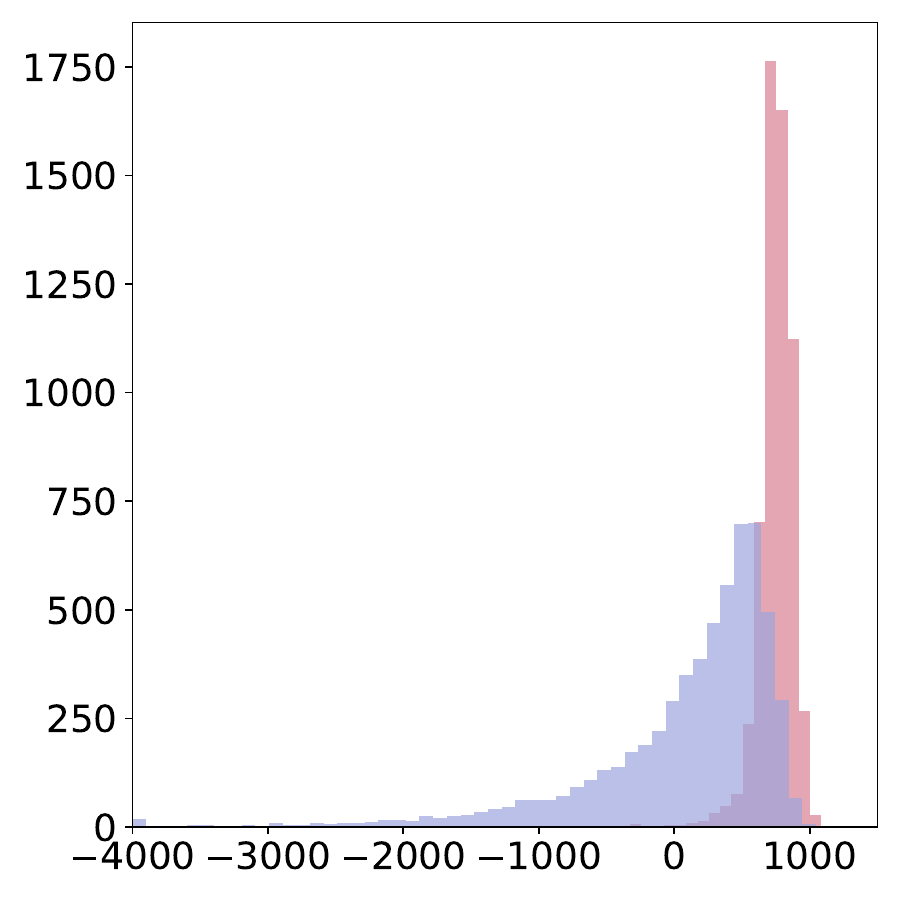}} \hspace{0.015\textwidth}
    \subfloat[\footnotesize BigGAN]{\includegraphics[width=0.12\textwidth]{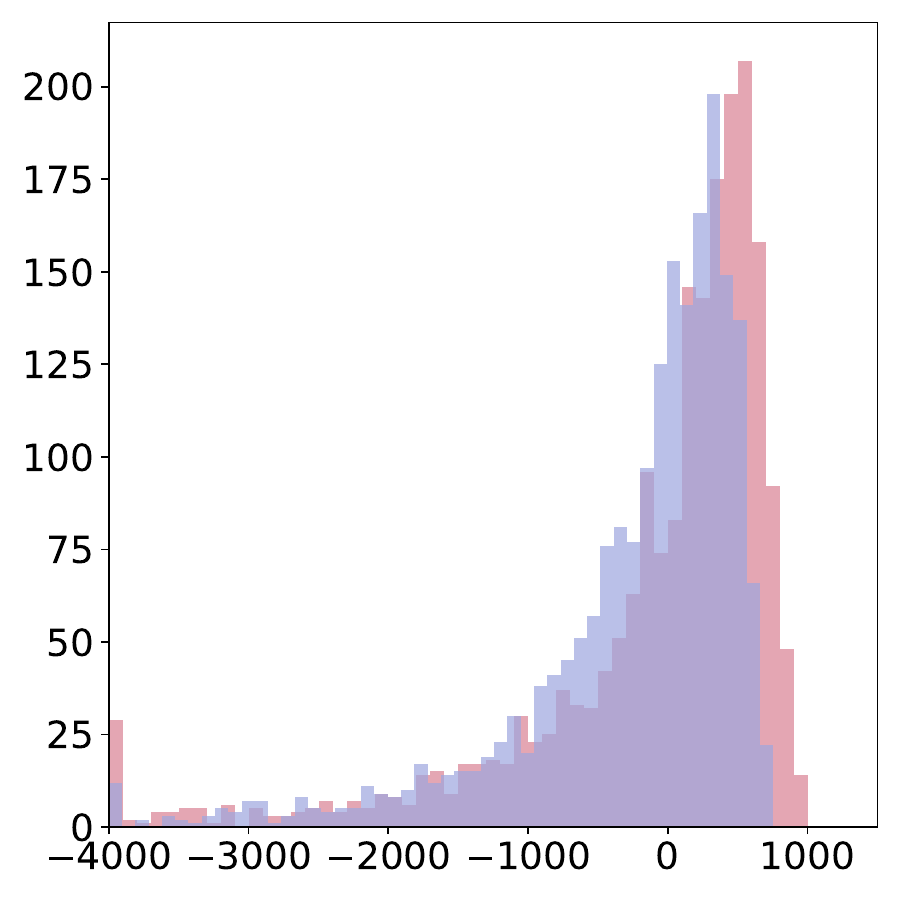}} \hspace{0.015\textwidth}
    \subfloat[\footnotesize CycleGAN]{\includegraphics[width=0.12\textwidth]{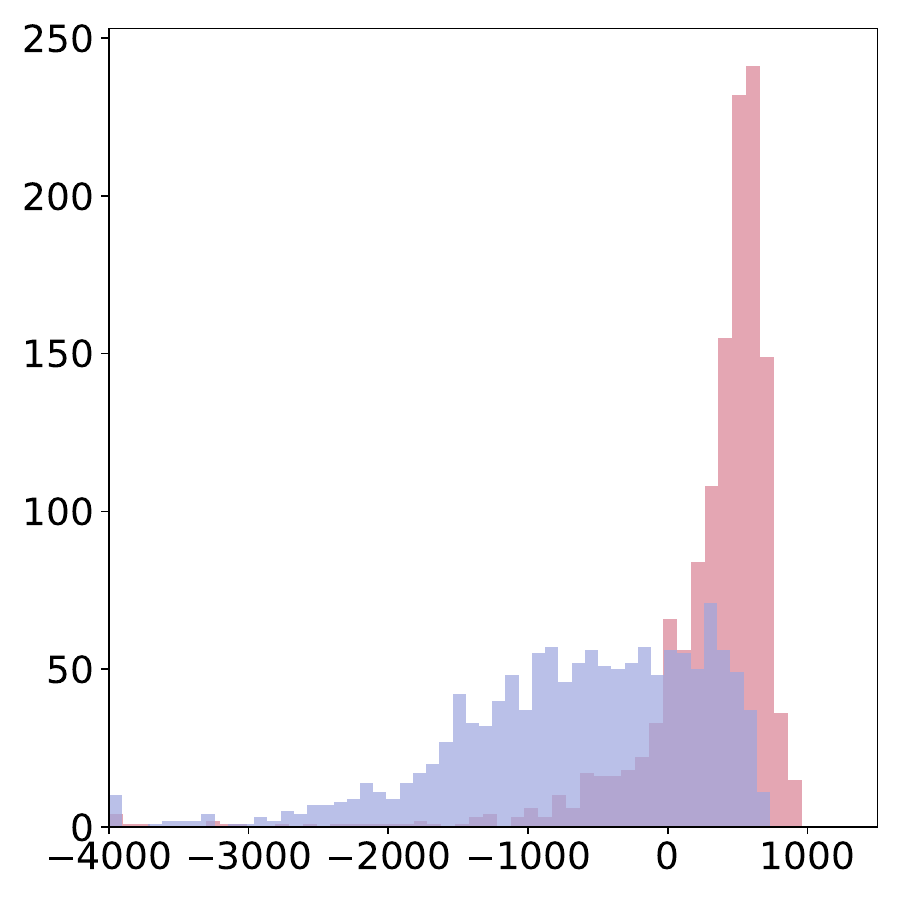}} \hspace{0.015\textwidth}
    \subfloat[\footnotesize StarGAN]{\includegraphics[width=0.12\textwidth]{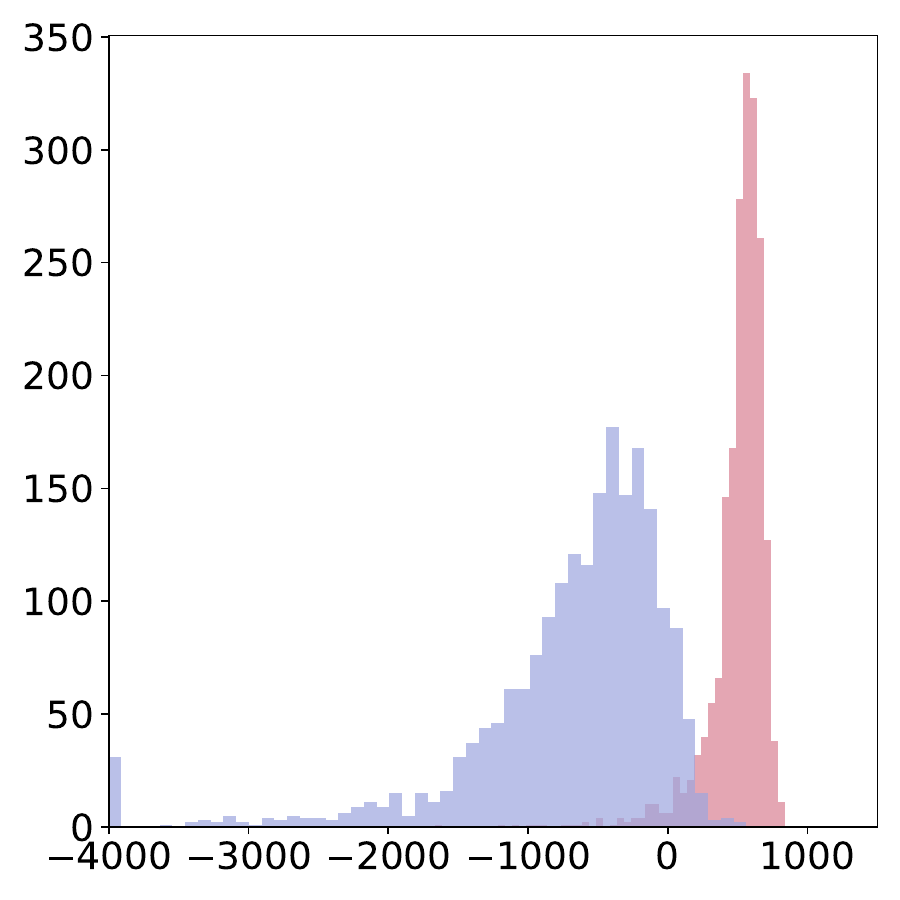}} \hspace{0.015\textwidth}
    \subfloat[\footnotesize GauGAN]{\includegraphics[width=0.12\textwidth]{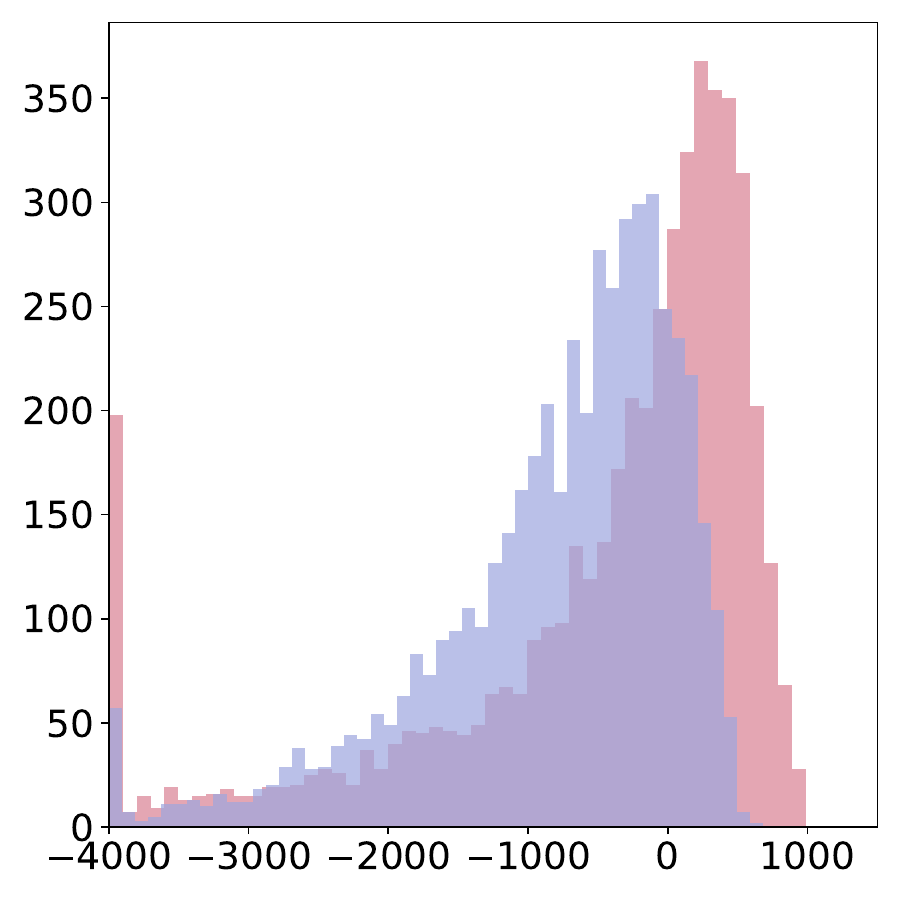}}
    \\[1pt]

    \subfloat[\footnotesize StyleGAN2]{\includegraphics[width=0.12\textwidth]{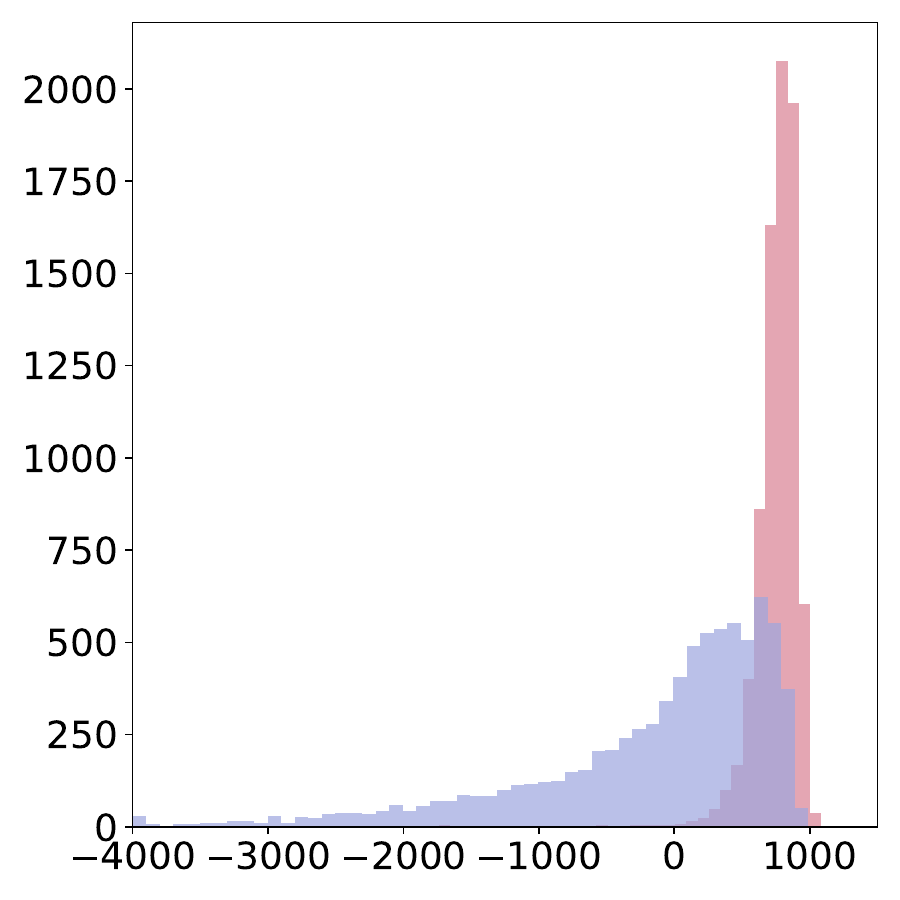}} \hspace{0.015\textwidth}
    \subfloat[\footnotesize WFIR]{\includegraphics[width=0.12\textwidth]{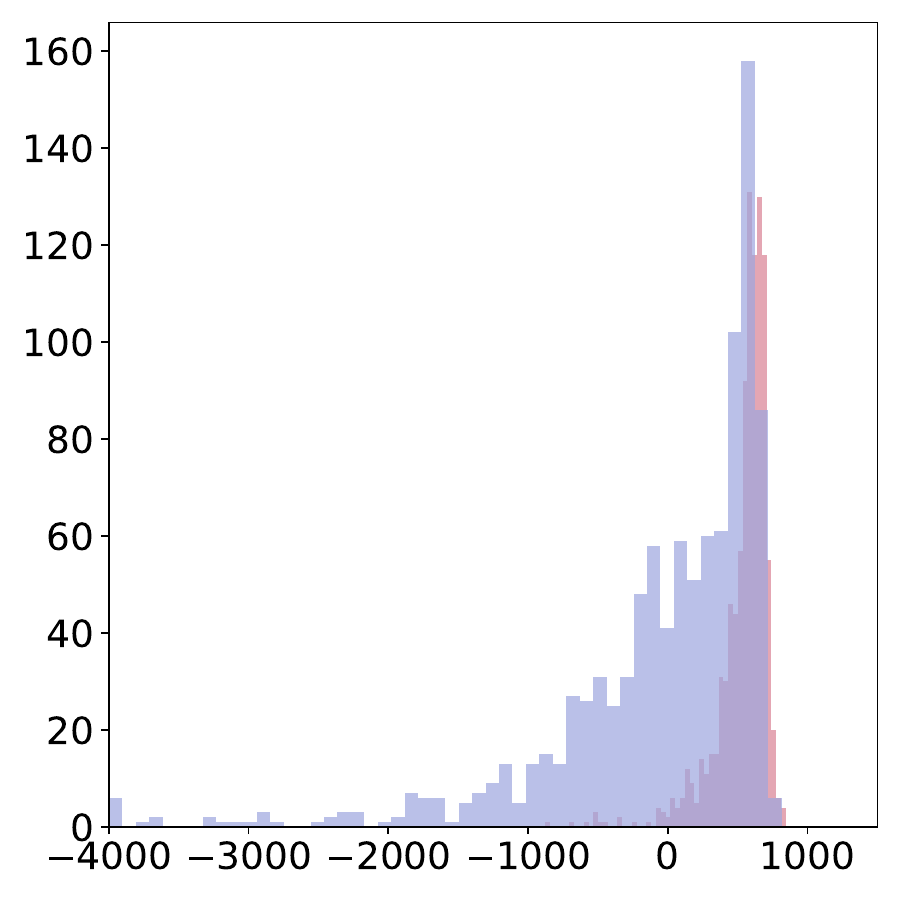}} \hspace{0.015\textwidth}
    \subfloat[\footnotesize ADM]{\includegraphics[width=0.12\textwidth]{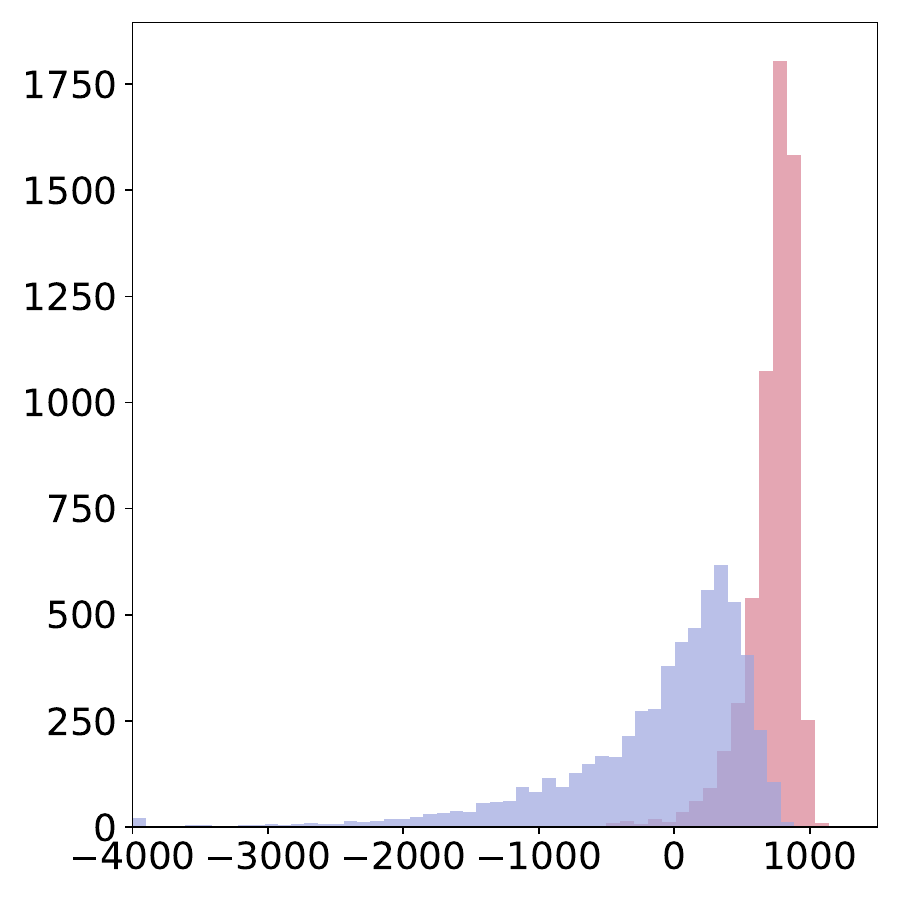}} \hspace{0.015\textwidth}
    \subfloat[\footnotesize Glide]{\includegraphics[width=0.12\textwidth]{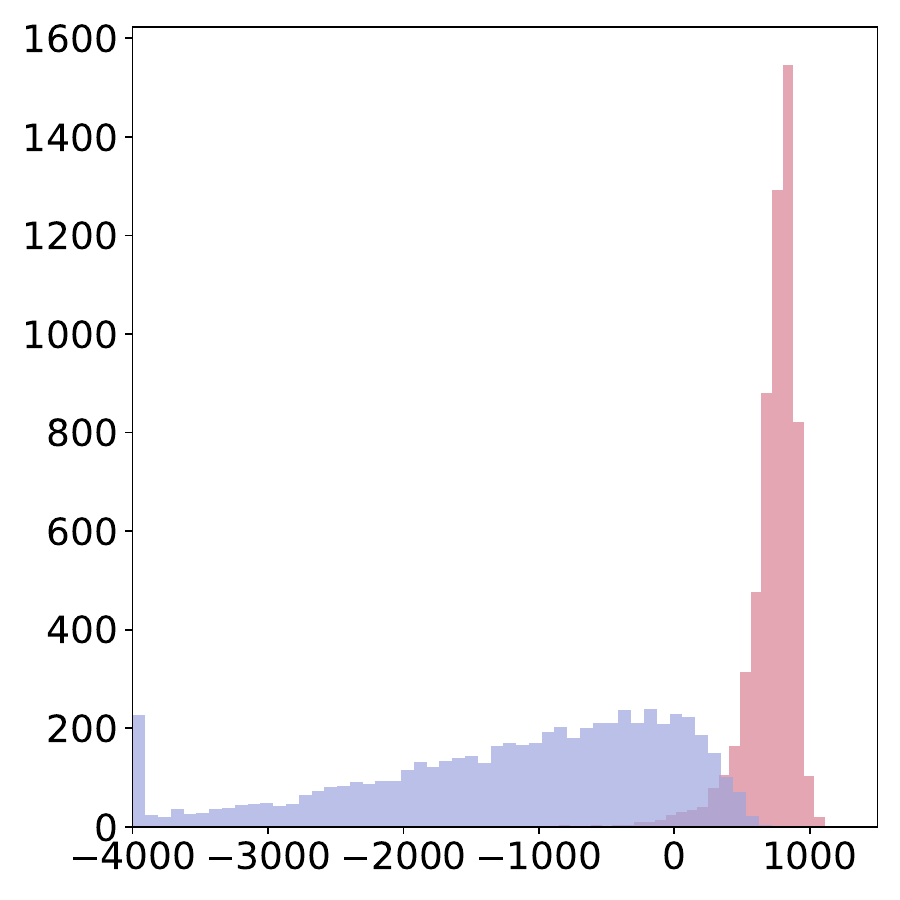}} \hspace{0.015\textwidth}
    \subfloat[\footnotesize Midjourney]{\includegraphics[width=0.12\textwidth]{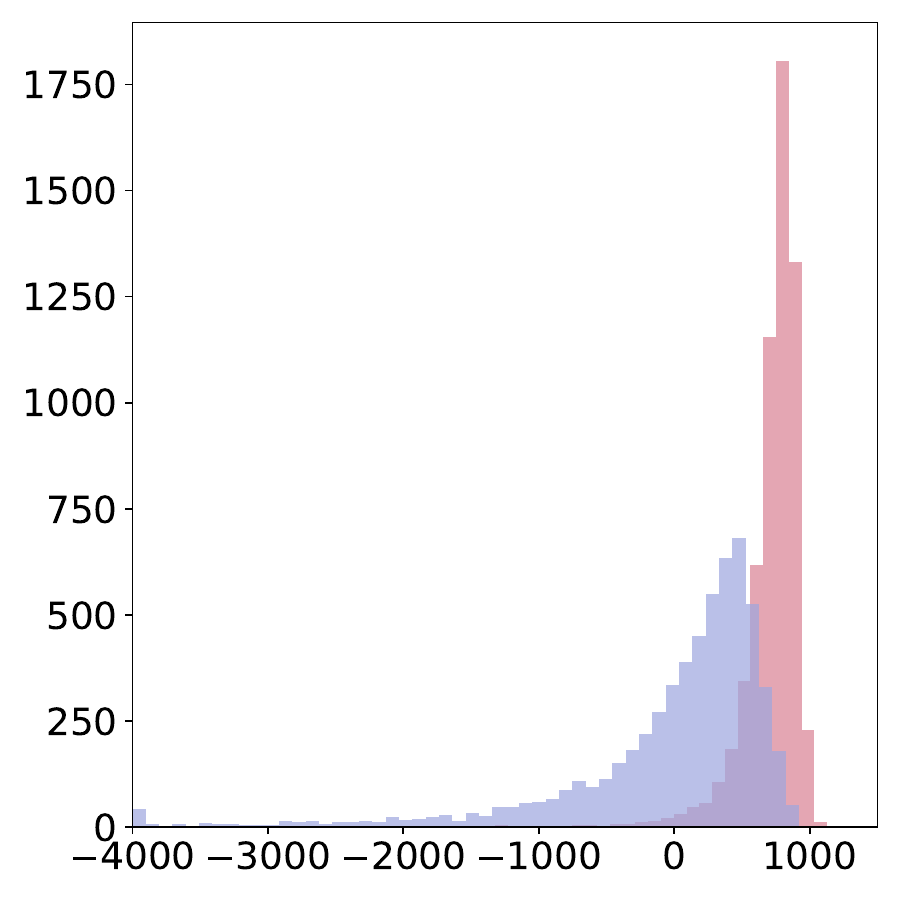}} \hspace{0.015\textwidth}
    \subfloat[\footnotesize SDv1.4]{\includegraphics[width=0.12\textwidth]{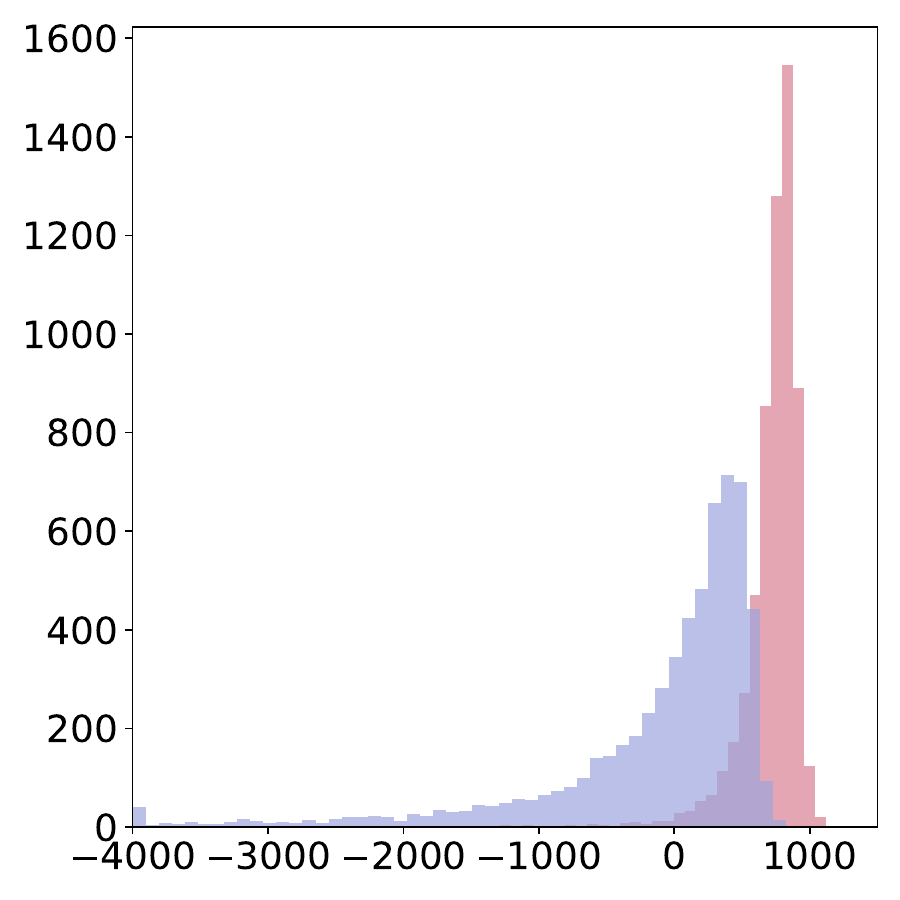}}
    \\[1pt]

    \hspace*{\fill}
    \subfloat[\footnotesize SDv1.5]{\includegraphics[width=0.12\textwidth]{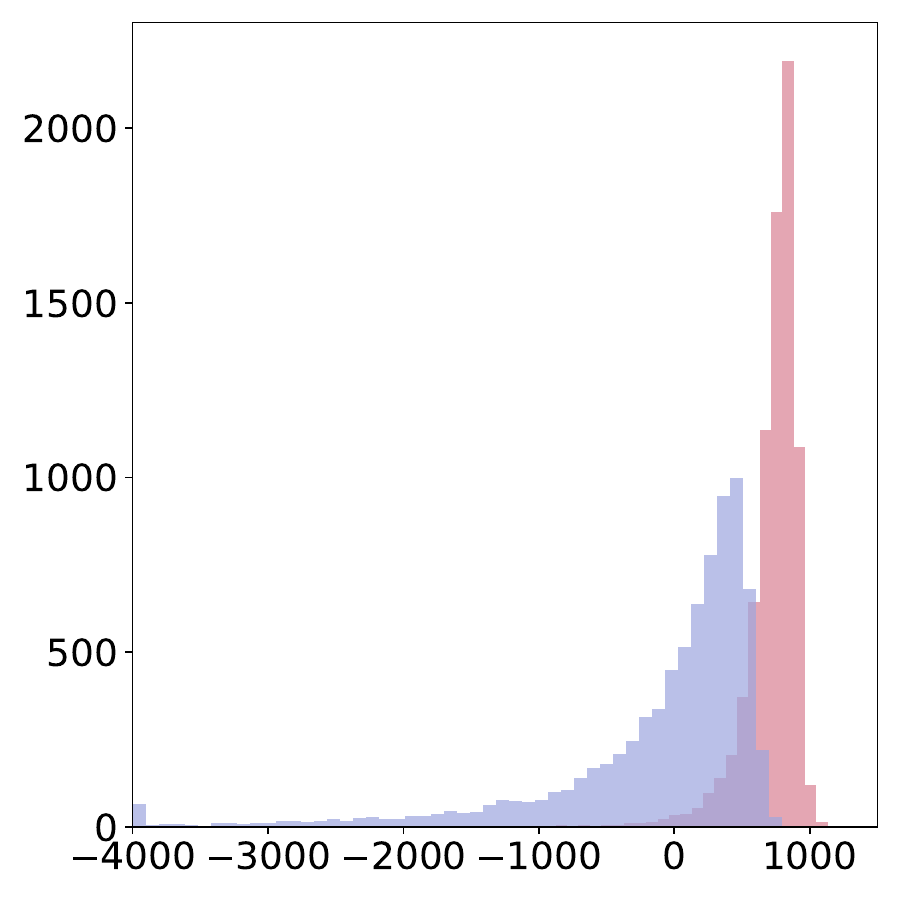}} \hspace{0.005\textwidth}
    \subfloat[\footnotesize VQDM]{\includegraphics[width=0.12\textwidth]{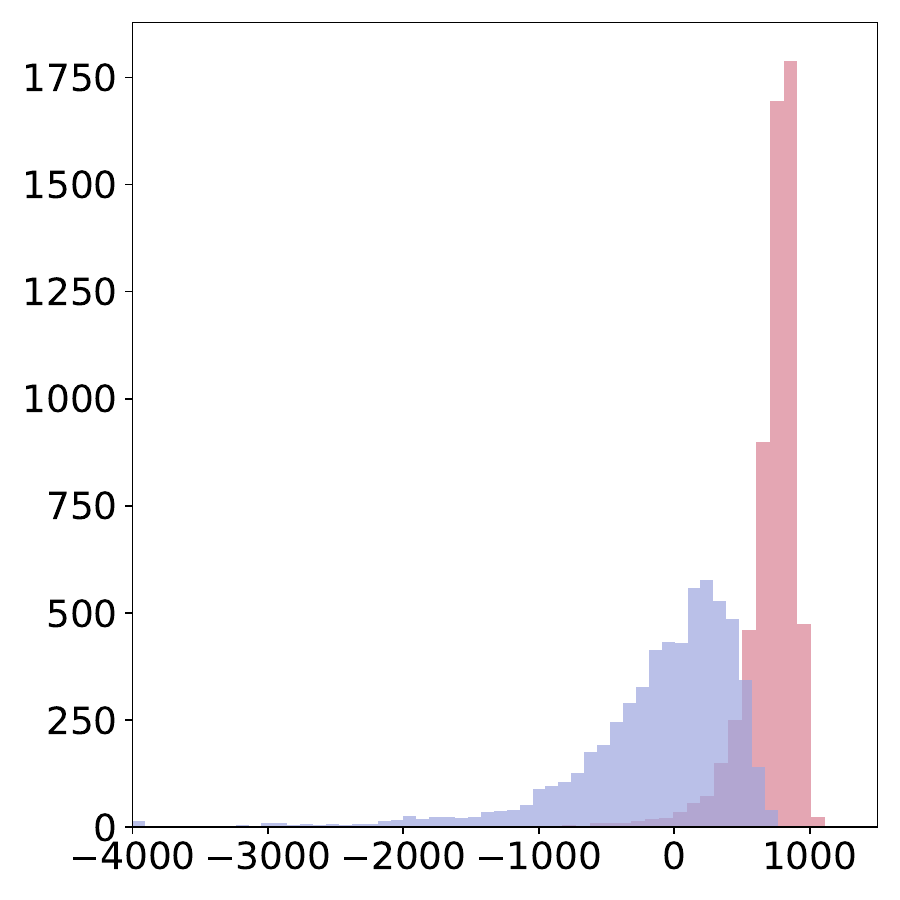}} \hspace{0.005\textwidth}
    \subfloat[\footnotesize WUKONG]{\includegraphics[width=0.12\textwidth]{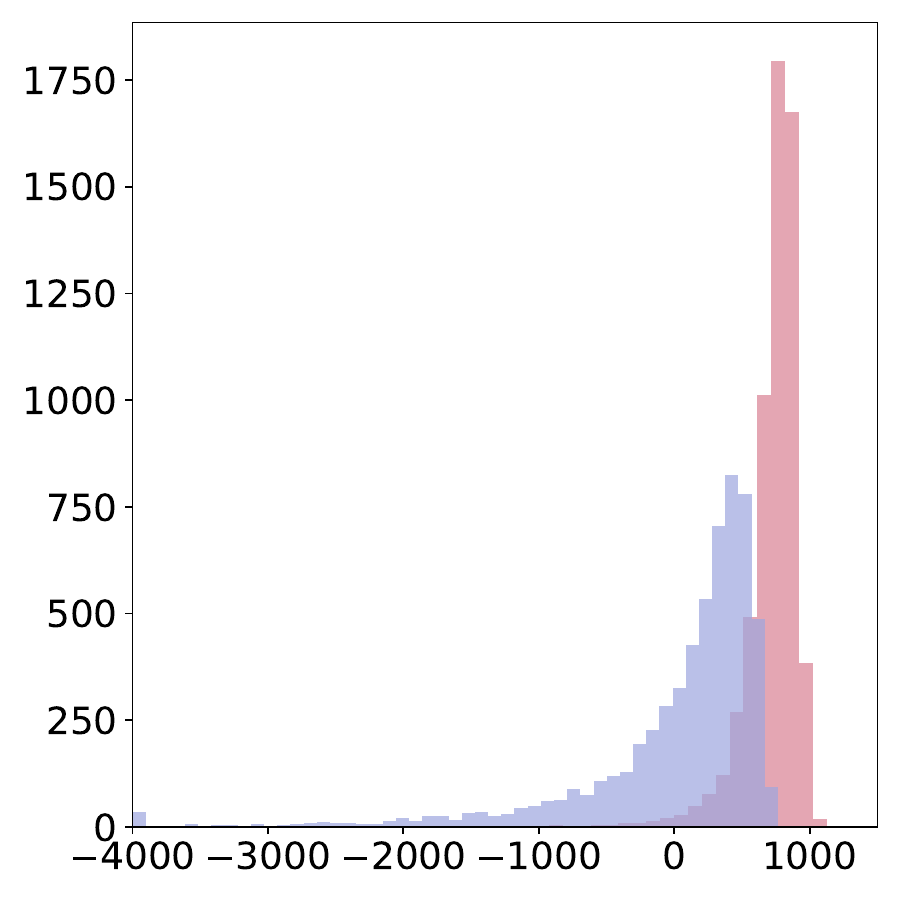}} \hspace{0.005\textwidth}
    \subfloat[\footnotesize DALLE2]{\includegraphics[width=0.12\textwidth]{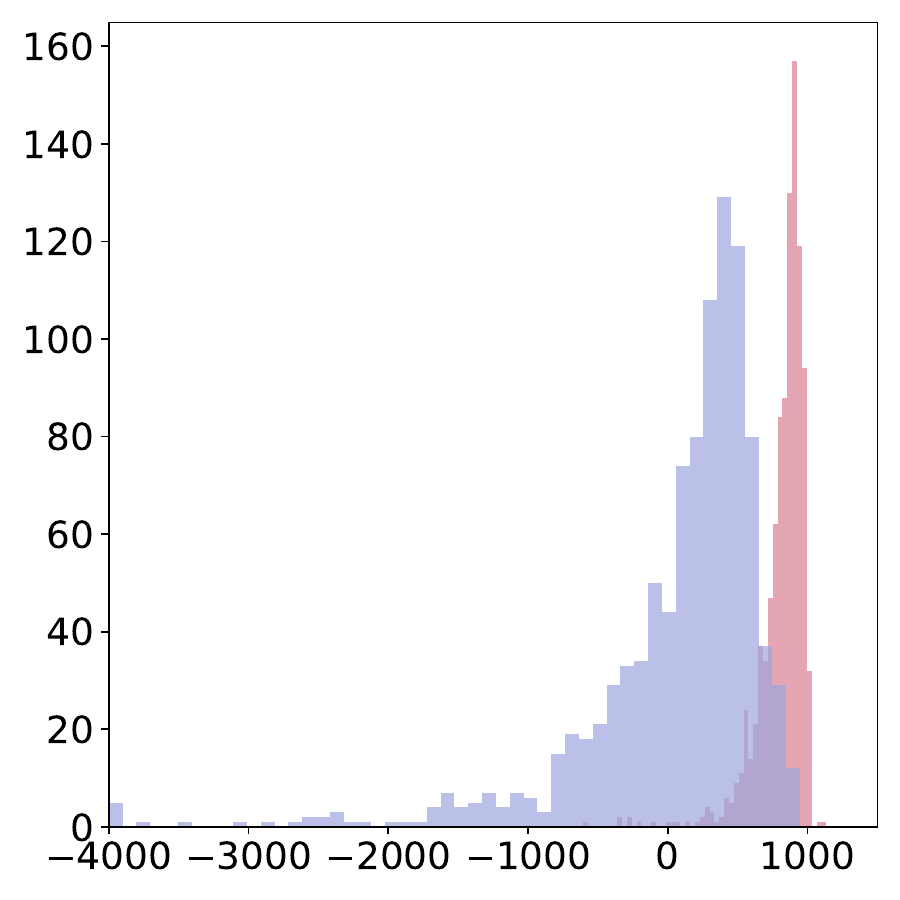}} \hspace{0.005\textwidth}
    \subfloat[\footnotesize SDXL]{\includegraphics[width=0.12\textwidth]{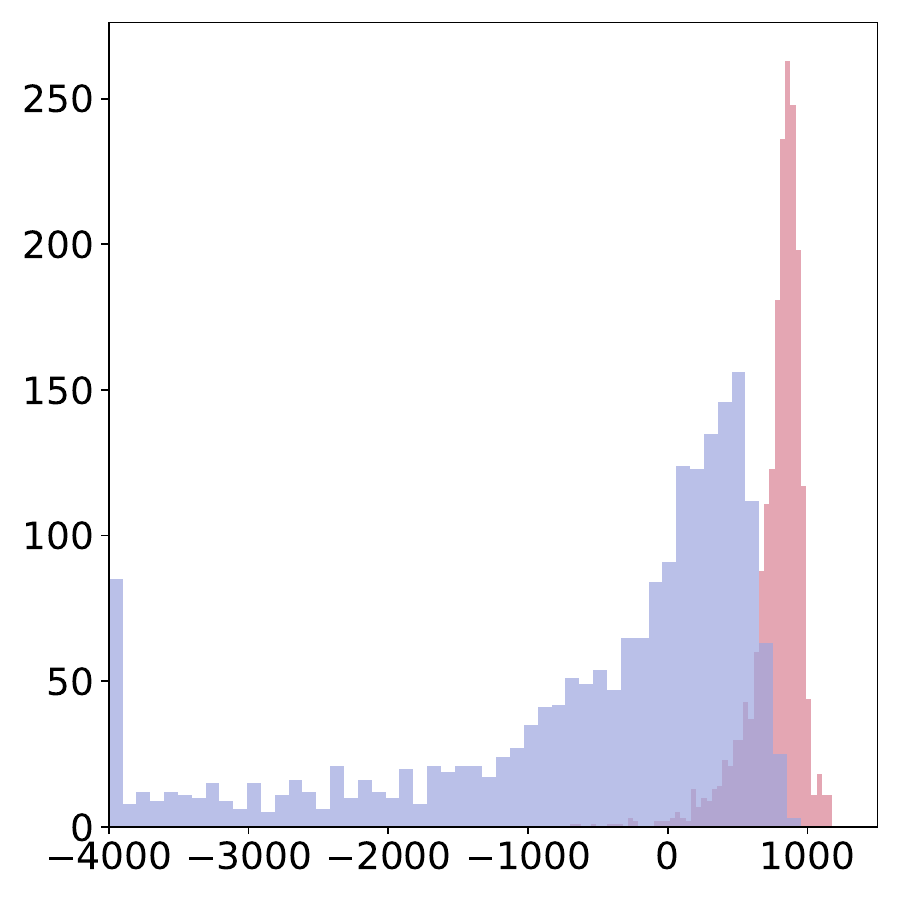}}
    \hspace*{\fill}
    
    \end{minipage}

    \caption{SDAIE log-likelihood profiles for photographic (red) and AI-generated (blue) images.}
    \label{fig_likelihood}
\end{figure*}

For a fair and reproducible comparison, we retrain all competing methods on the same training set using their official code releases. Importantly, we verify our reproduced results against the numbers reported in the corresponding papers to ensure consistency and reliability.

\parhead{Implementation Details.} We learn the EXIF-induced feature extractor exclusively from photographs. For each training image, we extract $N=16$ scrambled patches of size $S=64$. The feature extractor uses $\vert\mathcal{L}\vert=11$ convolution blocks with $3\times 3$ kernels, followed by covariance pooling to produce a $528$-dim image representation. The trade-off parameters ${\boldsymbol{\alpha}, \boldsymbol{\beta}}$ are all set to $1$. Optimization is carried out by Adam with a fixed learning rate of $10^{-4}$, a minibatch size of $64$, and a total of $30,000$ iterations.

For one-class SDAIE, we model the distribution of EXIF-induced photographic features with 
a GMM using $K=5$ components. At test time, the detection threshold $\tau$ is set to the $2$-nd quantile of training likelihoods, which corresponds to a training-set false alarm rate below $2\%$.
For binary variant SDAIE$^\dagger$, we attach a sigmoid classification head and optimize the sum of the cross-entropy and the representation-alignment regularizer (Eq.~\eqref{eq_binary_classification}). We use Adam with learning rate $10^{-3}$, minibatch size $100$, and $1,800$ iterations; the trade-off weight is $\gamma=0.05$.

 To improve generalization~\cite{wang2020cnn, ojha2023towards}, we treat common image processing operations as benign perturbations and apply them as data augmentations during training. Specifically, we include
\begin{itemize}
    \item JPEG compression with the quantization factor sampled from $\mathcal{U}[90,100]$;
    \item Bilinear downsampling with the scaling ratio sampled from $\mathcal{U}[0.25, 1]$;
    \item Gaussian blurring with the standard deviation sampled from $\mathcal{U}[0,1]$.
\end{itemize}
In the detection phase, we extract all non-overlapping $64\times 64$ patches from each test image, and feed them to the learned detectors to screen AI-generated content.

\subsection{Main Results}
We employ two complementary metrics to assess AI-generated image detection performance: 1) detection accuracy (Acc) and 2) mean average precision (mAP).

\begin{table*}[ht]
  \centering
  \caption{Detection performance (Acc/mAP) on GAN-based generators.}
  \label{tab_merged_main_gan}
  
  \resizebox{\textwidth}{!}{%
  \small
  \begin{tabular}{lccccccccc}
    
    \toprule
    Method & ProGAN & StyleGAN & BigGAN & CycleGAN & StarGAN & GauGAN & StyleGAN2 & WIFR & Average \\
    \midrule
    CNNSpot & \textbf{100.0}/\textbf{100.0} & 90.2/99.8  & 71.20/86.00  & 87.6/94.9  & 94.6/99.0  & 81.4/90.8  & 86.9/99.5  & \textbf{91.7}/\textbf{99.9} & 88.0/96.2\\
    GramNet & \textbf{100.0}/\textbf{100.0} & 87.1/99.2  & 67.30/81.80  & 86.1/95.3  & 95.1/99.2  & 69.4/85.0  & 87.3/99.1  & 86.8/95.2  & 84.9/94.4\\
    Frank20 & 99.4/\textbf{100.0} & 78.0/89.0  & 82.00/93.60  & 78.8/84.8  & 94.6/99.5  & 80.6/82.8  & 66.2/82.5  & 50.8/55.9  & 78.8/86.0\\
    Ju22 & \textbf{100.0}/\textbf{100.0} & 85.2/99.5  & 77.40/90.70  & 87.0/95.5  & 97.0/99.8  & 77.0/88.3  & 83.3/99.6  & 66.8/93.3  & 84.2/95.8\\
    LNP & \textbf{100.0}/\textbf{100.0} & 92.6/99.3  & 88.40/94.50  & 79.1/89.5  & \textbf{100.0}/\textbf{100.0} & 79.2/84.5  & 93.8/99.7  & 50.0/42.8  & 85.4/88.8\\
    LGrad & \textbf{100.0}/\textbf{100.0} & 90.5/98.9  & 88.80/96.30  & 85.7/94.0  & 99.6/\textbf{100.0} & 82.8/92.9  & 87.8/92.9  & 58.2/60.1  & 86.7/91.9\\
    DIRE-G & 95.2/99.1  & 83.0/91.7  & 70.10/75.30  & 74.2/80.6  & 95.5/99.3  & 67.8/72.2  & 75.3/88.3  & 58.1/60.1  & 77.4/83.3\\
    DIRE-D & 52.8/58.8  & 51.3/56.7  & 49.70/46.90  & 49.6/50.0  & 46.7/40.6  & 51.2/47.3  & 51.7/58.0  & 53.3/59.0  & 50.8/52.2\\
    UnivFD & 99.8/\textbf{100.0} & 84.9/97.6  & \textbf{95.10}/\textbf{99.30} & \textbf{98.3}/\textbf{99.8} & 95.8/99.4  & \textbf{99.5}/\textbf{100.0} & 75.0/97.9  & 86.9/96.7  & 91.9/\textbf{98.8}\\
    NPR & 99.9/\textbf{100.0} & 96.1/99.9  & 87.30/94.00  & 90.3/99.1  & 99.6/\textbf{100.0} & 85.4/88.7  & 98.1/\textbf{100.0} & 60.7/94.0  & 89.7/97.0\\
    \midrule
    SDAIE & 78.6/87.3  & 84.8/89.7  & 57.90/66.00  & 66.2/85.9  & 74.5/99.1  & 55.3/73.4  & 85.0/89.5  & 71.5/79.4  & 71.7/83.8\\
    SDAIE$^\dagger$ & \textbf{100.0}/\textbf{100.0} & \textbf{99.8}/\textbf{100.0} & 90.50/97.30  & 91.9/99.4  & \textbf{100.0}/\textbf{100.0} & 84.0/89.9  & \textbf{99.2}/\textbf{100.0} & 84.5/98.1  & \textbf{93.7}/98.1\\

    \bottomrule
  \end{tabular}%
  }
\end{table*}

\begin{table*}[ht]
  \centering
  \caption{Detection performance (Acc/mAP) on diffusion-based generators.}
  \label{tab_merged_main_diffusion}
  \resizebox{\textwidth}{!}{%
  \small
  \begin{tabular}{lcccccccccc}
    \toprule
    Method & ADM & Glide & Midjourney & SDv1.4 & SDv1.5 & VQDM & WUKONG & DALLE2 & SDXL & Average \\
    \midrule
    CNNSpot & 60.4/75.7  & 58.1/72.3  & 51.4/66.2  & 50.6/61.2  & 50.5/61.6  & 56.5/68.8  & 51.0/57.3  & 50.5/53.5  & 53.0/72.6  & 53.6/65.5\\
    GramNet & 58.6/73.1  & 54.5/66.8  & 50.0/56.8  & 51.7/59.8  & 52.2/60.4  & 52.9/61.1  & 50.8/55.6  & 49.3/49.8  & 64.5/68.2  & 53.8/61.3\\
    Frank20 & 63.4/61.8  & 54.1/52.9  & 45.9/46.1  & 38.8/37.8  & 39.2/37.8  & 77.8/85.1  & 40.3/39.6  & 34.7/38.2  & 51.2/49.5  & 49.5/49.9\\
    Ju22 & 49.0/94.1  & 57.2/77.5  & 52.2/70.0  & 51.0/65.4  & 51.4/65.7  & 55.1/75.6  & 51.7/64.6  & 52.8/68.1  & 55.6/79.4  & 52.9/73.4\\
    LNP & 83.9/93.4  & 83.5/92.8  & 69.6/86.9  & 89.3/96.3  & 88.8/96.0  & 85.0/94.9  & 86.4/95.3  & 92.5/98.3  & 87.8/87.8  & 85.2/93.5\\
    LGrad & 66.2/73.5  & 71.7/84.4  & 70.5/77.7  & 65.2/68.9  & 65.9/69.3  & 74.7/79.5  & 60.3/65.7  & 71.3/86.4  & 71.3/80.0  & 68.6/76.2\\
    DIRE-G & 75.8/85.8  & 71.8/78.4  & 58.0/61.9  & 49.7/49.9  & 49.8/49.5  & 53.7/54.6  & 54.5/55.4  & 66.5/74.5  & 55.4/54.0  & 59.5/62.7\\
    DIRE-D & \textbf{98.3}/\textbf{99.8} & 92.4/99.5  & 89.5/97.3  & 91.2/98.6  & 91.6/98.8  & \textbf{91.9}/99.0  & 90.9/98.4  & 92.5/99.7  & 91.3/99.1  & 92.2/98.9\\
    UnivFD & 66.9/86.8  & 62.5/83.8  & 56.1/74.0  & 63.7/86.1  & 63.5/85.8  & 85.3/96.5  & 70.9/91.1  & 50.8/63.0  & 50.7/67.6  & 63.4/81.6\\
    NPR & 84.9/98.3  & \textbf{96.7}/\textbf{99.6} & 92.6/99.0  & \textbf{97.4}/99.7  & \textbf{97.5}/99.6  & 90.1/\textbf{99.2} & 91.7/98.9  & \textbf{99.6}/\textbf{100.0} & \textbf{98.6}/99.6  & 94.3/\textbf{99.3}\\
    \midrule
    SDAIE & 88.9/95.9  & 93.2/99.3  & 85.6/92.8  & 89.7/96.5  & 89.7/96.4  & 90.8/97.2  & 87.5/95.6  & 89.0/95.9  & 89.4/96.6  & 89.3/96.2\\
    SDAIE$^\dagger$ & 92.9/98.3  & 93.1/98.5  & \textbf{95.4}/\textbf{99.4} & 95.8/\textbf{99.9} & 95.7/\textbf{99.8} & 91.5/97.9  & \textbf{95.0}/\textbf{99.4} & 97.1/99.9  & 96.6/\textbf{100.0} & \textbf{94.8}/99.2\\
    \bottomrule
  \end{tabular}%
  }
\end{table*}

\parhead{One-Class SDAIE Results.} We first conduct a qualitative assessment of SDAIE. Fig.~\ref{fig_tsne} applies t-SNE~\cite{van2008visualizing} to the EXIF-induced features, and reveals a clear separation between photographic and AI-generated images. This separation shows that the learned representations effectively capture camera-intrinsic regularities, supporting the feasibility of detecting AI-generated images using self-supervised learning solely on photographic data. Fig.~\ref{fig_likelihood} further shows the log-likelihood histograms, where AI-generated images consistently attain lower log-likelihoods than photographic ones in most cases, with only minor distributional overlap. Such behavior is precisely what is desired in a one-class setting, as it allows for reliable discrimination of AI-generated images while maintaining a low false alarm rate.
 
\begin{figure}[]
    \centering
    \includegraphics[width=\linewidth]{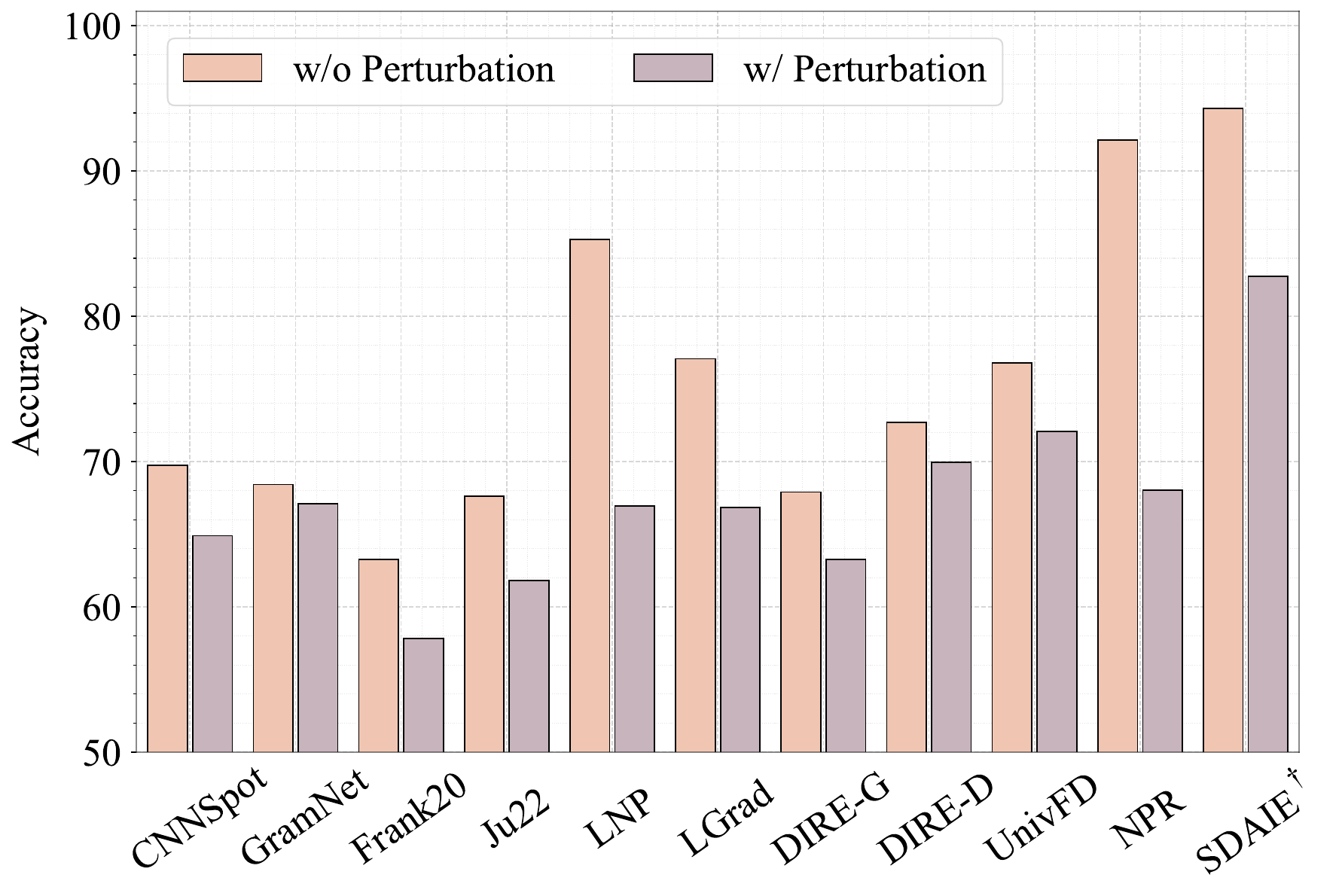}
    \caption{Overall robustness of AI-generated image detectors under benign post-processing perturbations.}
    \label{fig_comprehensive_comparison}
\end{figure}

We next report quantitative results in Tables~\ref{tab_merged_main_gan} and~\ref{tab_merged_main_diffusion}. The seventeen evaluated generative models are grouped into two categories: GAN-based and diffusion-based generators. Across this diverse collection of models, SDAIE attains non-trivial detection performance despite never observing AI-generated images during training. This outcome underscores the effectiveness of our EXIF-induced self-supervised learning strategy in shaping a feature space where AI-generated images emerge as low-likelihood outliers relative to the distribution of photographic images.

Interestingly, SDAIE attains noticeably higher mean detection performance on diffusion-based generators than on GAN-based ones. This may arise because EXIF-induced representations are explicitly aligned with camera-intrinsic residual statistics. Diffusion models synthesize images via iterative denoising, without modeling the physical imaging pipeline, and consequently fail to reproduce sensor noise, demosaicing periodicity, and compression fingerprints captured by SDAIE. By contrast, several of the GAN-based benchmarks operate on narrow, constrained photographic domains and can partially mimic camera-like residuals, leading to greater overlap with the photographic manifold and slightly reduced one-class separability.

\parhead{Binary SDAIE$^\dagger$ Results.} We next evaluate the binary variant SDAIE$^\dagger$, which is regularized by the EXIF-induced feature extractor. Tables~\ref{tab_merged_main_gan} and~\ref{tab_merged_main_diffusion} show that SDAIE$^\dagger$ surpasses most competing methods by a clear margin across both GAN-based and diffusion-based generators. All competing detectors, except DIRE-D, are trained on a mixture of photographic and ProGAN-generated images, yet most degrade substantially when evaluated on diffusion-based content, whereas SDAIE$^\dagger$ maintains strong accuracy in this setting. We attribute this superior cross-model generalization to the core design of SDAIE$^\dagger$, which detects AI-generated images by leveraging EXIF-induced representations intrinsic to photographic images rather than artifacts tied to specific generative models. Consequently, even when ProGAN is the only synthetic source during training, SDAIE$^\dagger$ remains effective on diffusion-generated images.

In contrast, conventional methods are explicitly designed and optimized to capture ``universal'' generative artifacts, whose discriminative power inevitably diminishes as generation architectures and synthesis pipelines evolve. DIRE-D, which is trained on diffusion-based synthetic data, exemplifies this limitation: it achieves competitive detection performance on diffusion-generated images but suffers pronounced accuracy degradation on GAN-generated images due to mismatch in 
artifact characteristics. Among the 
remaining, UnivFD is the closest competitor to SDAIE$^\dagger$ on GAN-based generators. Yet, its semantics-oriented detection principle leads to strong performance mainly where GAN-generated images exhibit more conspicuous semantic or structural anomalies, and to weaker performance on diffusion-based generators whose outputs are often semantically coherent and visually refined.
 Compared with another state-of-the-art method, NPR \cite{liu2022detecting}, SDAIE$^\dagger$ achieves higher accuracy, especially on GAN-based generators. NPR identifies AI-generated images via artifacts introduced by upsampling layers, and its generalization relies on the ubiquity and stability of such operators in image generation. However, as modern generators increasingly adopt architectures with reduced or carefully compensated upsampling, these artifact cues become unreliable or even absent in practice. In addition, camera pipelines may introduce aliasing and interpolation effects that resemble synthetic upsampling, leading to inflated false negatives. 
 
 \begin{figure*}[htbp]
    \centering
    \captionsetup[subfloat]{labelformat=empty}
    \subfloat[JPEG compression]{%
        \includegraphics[width=0.33\textwidth]{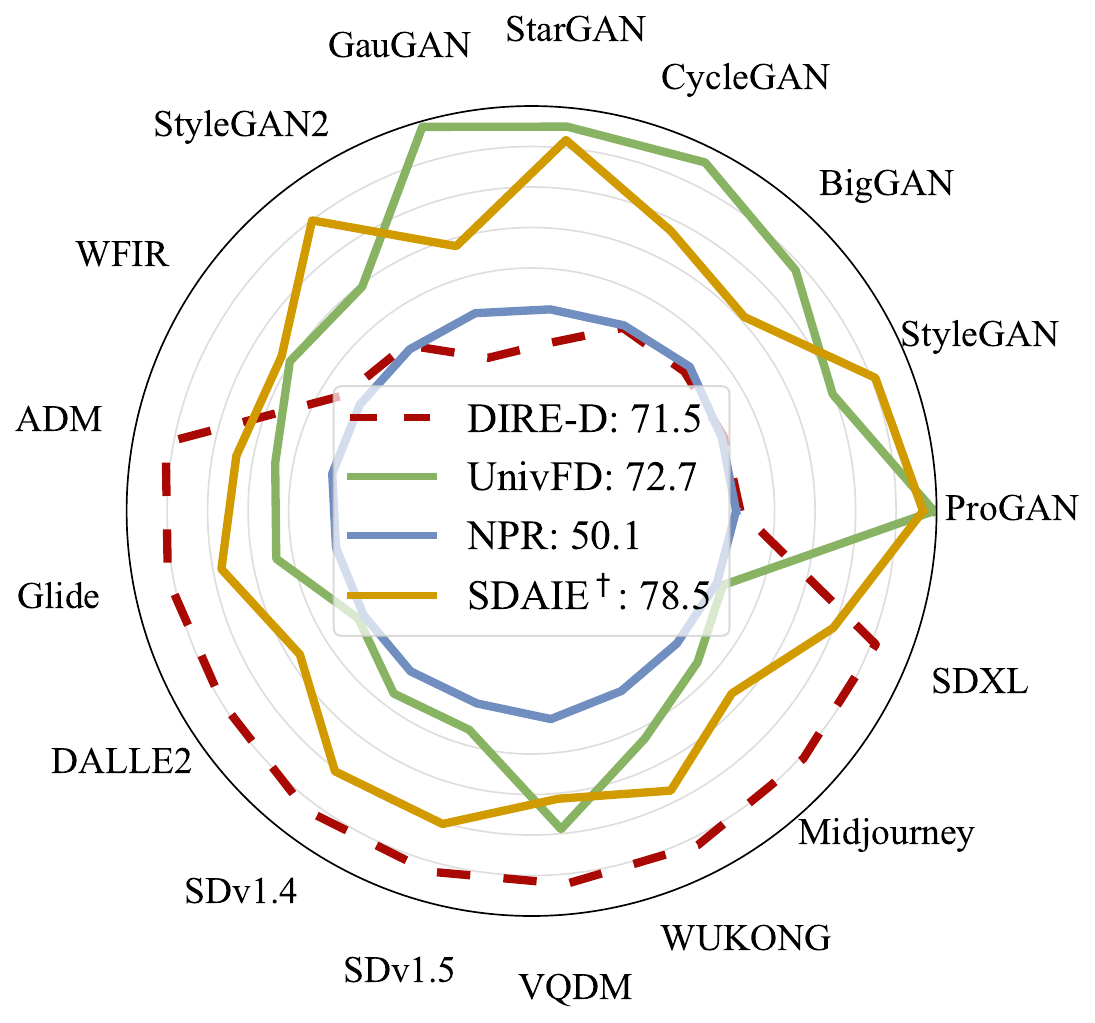}%
    }\hfill
    \subfloat[Gaussian blurring]{%
        \includegraphics[width=0.33\textwidth]{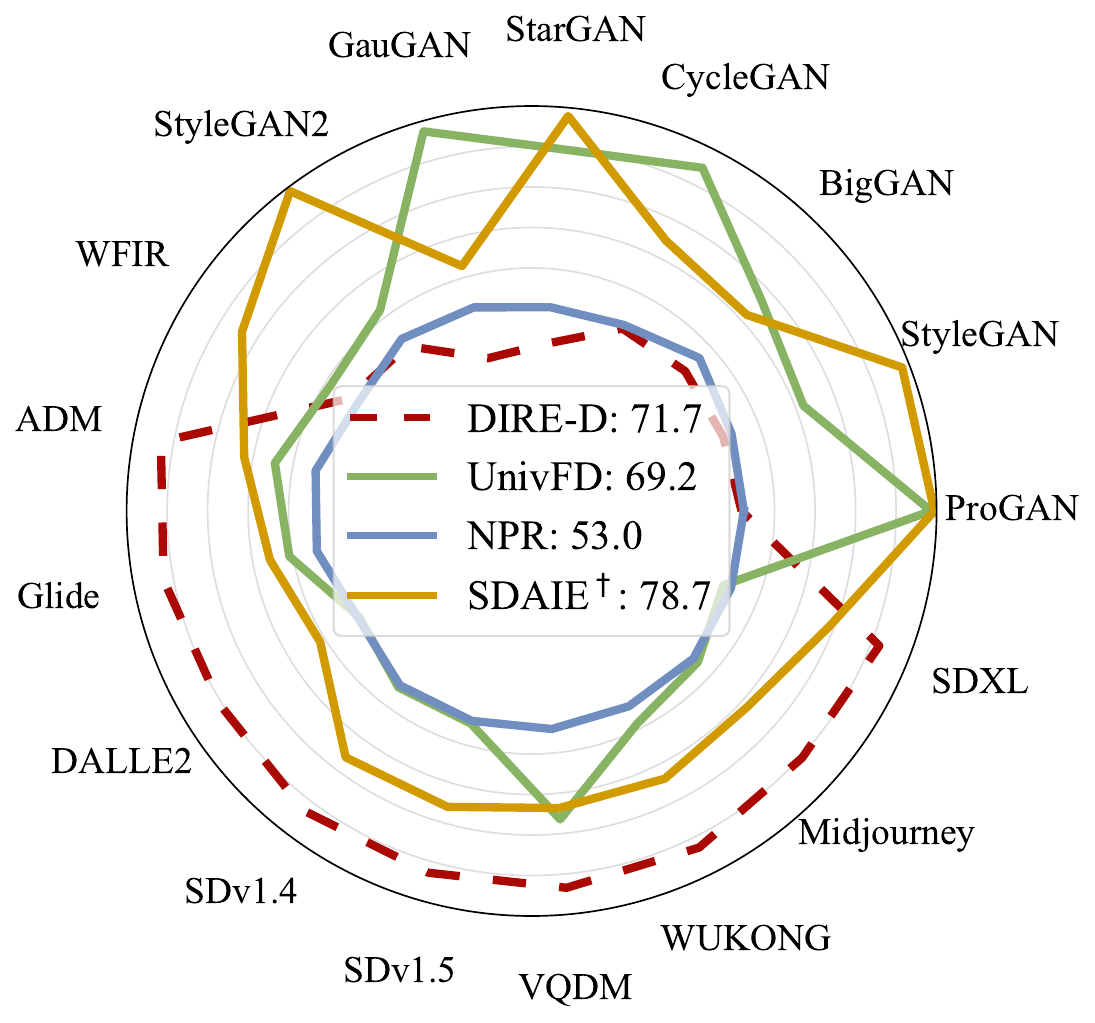}%
    }\hfill
    \subfloat[Downsampling]{%
        \includegraphics[width=0.33\textwidth]{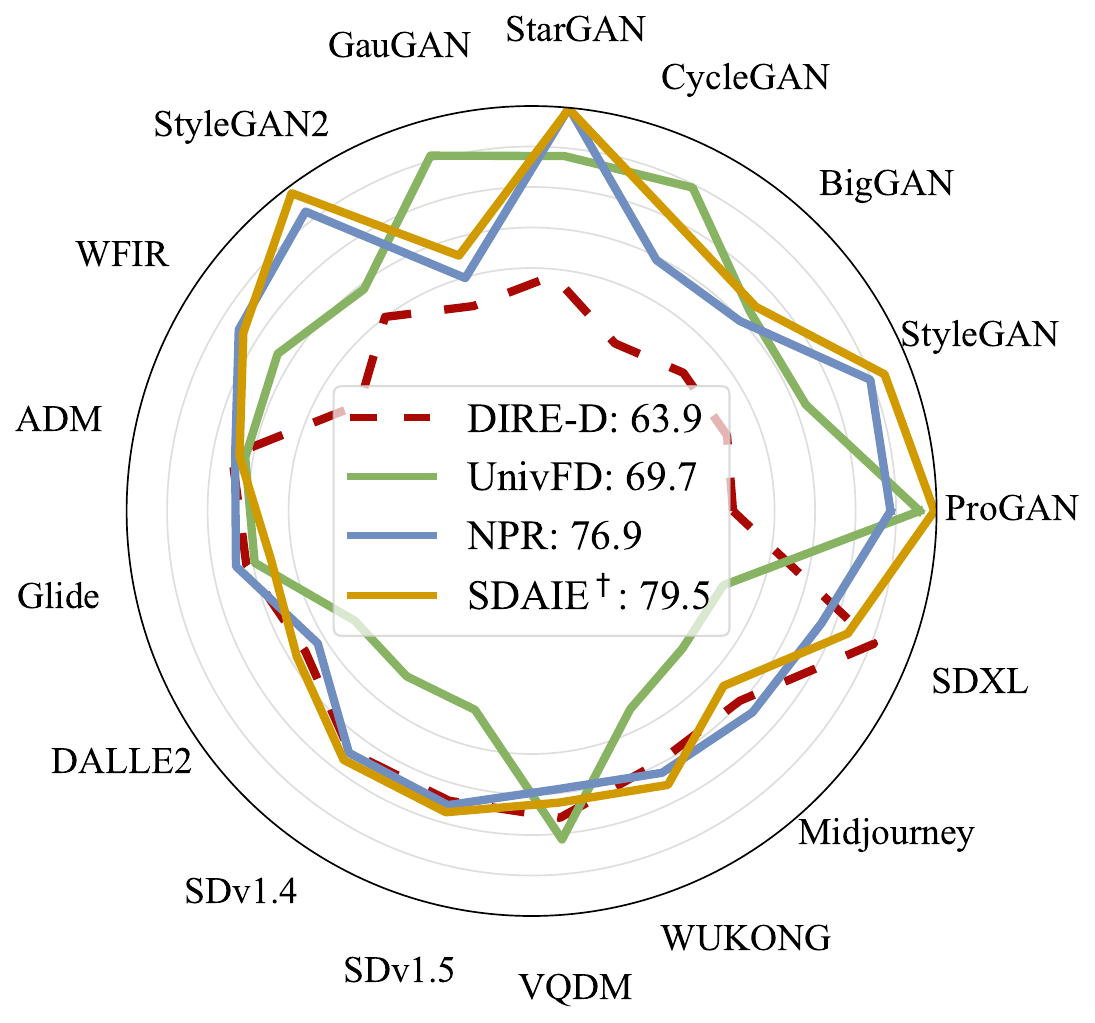}%
    }
    \caption{Robustness of AI-generated image detectors to individual benign post-processing operators. The legend represents the average detection accuracy of each detector.}
    \label{fig_robustness}
\end{figure*}

\parhead{Robustness to Benign Perturbations.} 
In real-world deployments, AI-generated images are seldom encountered in their pristine form; instead, they typically undergo routine post-processing operations such as JPEG compression, mild blurring, or rescaling along acquisition, transmission, and display pipelines. These label-preserving transformations may weaken or even obscure forensic cues, thereby challenging the robustness of detectors. To systematically assess this aspect, we subject all detectors to three commonly used operators:
 JPEG compression with a quality factor of $95$, Gaussian blurring with a standard deviation of $1$, and downsampling by a factor of $2$ along both spatial dimensions. Fig.~\ref{fig_comprehensive_comparison} summarizes the overall robustness by reporting, for each detector, the mean detection accuracy across seventeen generators. While several competing methods, especially NPR, perform strongly on pristine images, their mean accuracies drop substantially when benign perturbations are introduced. In contrast, SDAIE$^\dagger$ maintains high accuracy across all scenarios and achieves the best aggregated performance, suggesting that the EXIF-induced, camera-intrinsic representations affords a more stable basis for AI-generated image detection.

 \begin{table*}[]
\centering
\caption{Generalization (Acc) of AI-generated image detectors to emerging generators and in-the-wild images.}
\begin{tabular}{lcccccccc}
\toprule
Method          & FLUX.1 & FLUX.1-Kontext & SD-3.5-Turbo & Qwen-Image & Midjourney & Tongyi & Lummi & Average \\
\midrule
UnivFD          & 0.1  & 0.8  & 2.2  & 6.5  & 2.8  & 0.9  & 4.5  & 2.5 \\
NPR             & 90.0 & 90.0 & 89.8 & 70.0 & 98.8 & 81.1 & 69.9 & 84.2 \\
\midrule
SDAIE$^\dagger$ & \textbf{96.2} & \textbf{96.6} & \textbf{94.8} & \textbf{94.1} & \textbf{99.0} & \textbf{98.1} & \textbf{84.7} & \textbf{94.8} \\
\bottomrule
\end{tabular}
\label{tab_wild}
\end{table*}

 Fig.~\ref{fig_robustness} provides a more fine-grained view of the robustness of SDAIE$^\dagger$ against three strong competitors---DIRE-D, UnivFD, and NPR---by breaking down performance according to the specific post-processing operator. Generally, JPEG compression and Gaussian blurring emerge as the most challenging perturbations: both aggressively attenuate high-frequency components where many synthesis-related cues (\eg, periodic artifacts and subtle noise inconsistencies) reside. In contrast, $2\times$ downsampling tends to be comparatively less destructive, since it primarily reduces spatial resolution while still preserving a portion of the global structural and color statistics that remain informative for detection. SDAIE$^\dagger$ preserves its advantage under all three operators, showing the smallest relative degradation when strong high-frequency suppression is applied. This operator-wise analysis reveals that robustness to frequency-suppressing transformations is the key bottleneck for current detectors, and that the EXIF-induced, camera-intrinsic features offer a more resilient solution in this regime.

\parhead{Generalization to Emerging Generators and In-the-Wild Images.} Beyond the curated benchmark suite, we further assess whether SDAIE$^\dagger$ can keep pace with rapidly evolving generative models and in-the-wild content circulating on social media. To this end, we collect additional AI-generated images using four recent text-to-image models: FLUX.1~\cite{flux}, FLUX.1-Kontext~\cite{flux}, SD-3.5-Turbo~\cite{sd3_5}, and Qwen-Image~\cite{wu2025qwen}, all conditioned on captions from the COCO dataset~\cite{COCO}. We produce $1,000$ images per model. In parallel, we collect in-the-wild AI-generated images attributed to Midjourney~\cite{midjourney} (latest v7 release), Tongyi~\cite{tongyi}, and Lummi~\cite{Lummi}, each comprising $1,000$ images, from public social platforms. These images better reflect realistic use cases with prompt, style, and post-processing diversities.

\begin{table}[t]
  \centering
  \caption{Choice of feature extractor on detection performance (Acc/mAP).}
\begin{tabular}{lccc}
\midrule
Generator & CLIP & EAL & SDAIE \\
\midrule
ProGAN & 76.21/82.85 & 50.15/51.27 & \textbf{78.59}/\textbf{87.25} \\
StyleGAN & 51.99/50.18 & 48.61/68.26 & \textbf{84.84}/\textbf{89.73} \\
BigGAN & \textbf{71.03}/\textbf{73.71} & 33.80/38.20 & 57.85/66.01 \\
CycleGAN & \textbf{78.84}/\textbf{89.65} & 57.08/70.79 & 66.24/85.88 \\
StarGAN & 50.08/98.87 & 44.62/31.76 & \textbf{74.49}/\textbf{99.13} \\
GauGAN & \textbf{63.12}/70.25 & 41.06/39.05 & 55.33/\textbf{73.43} \\
StyleGAN2 & 37.07/43.36 & 46.35/55.28 & \textbf{85.02}/\textbf{89.51} \\
WFIR & 53.20/60.12 & 47.00/50.53 & \textbf{71.45}/\textbf{79.38} \\
ADM & 48.08/57.45 & 72.90/93.87 & \textbf{88.85}/\textbf{95.91} \\
Glide & 46.50/53.87 & \textbf{93.25}/97.96 & 93.24/\textbf{99.33} \\
Midjourney & 49.39/54.16 & 59.06/75.26 & \textbf{85.58}/\textbf{92.83} \\
SDv1.4 & 45.18/45.39 & 51.93/80.97 & \textbf{89.72}/\textbf{96.46} \\
SDv1.5 & 45.37/45.53 & 52.74/81.61 & \textbf{89.65}/\textbf{96.44} \\
VQDM & 45.58/53.47 & 67.06/91.94 & \textbf{90.76}/\textbf{97.19} \\
WUKONG & 49.33/48.94 & 53.87/81.63 & \textbf{87.49}/\textbf{95.59} \\
DALLE2 & 42.90/45.50 & 76.25/\textbf{99.00} & \textbf{88.95}/95.94 \\
SDXL & 51.85/53.99 & 53.35/82.45 & \textbf{89.35}/\textbf{96.59} \\
\midrule
Average & 53.28/60.43 & 55.83/69.99 & \textbf{81.02}/\textbf{90.39} \\
\bottomrule
\end{tabular}
  \label{tab_ablation_pretext}%
\end{table}%

Table~\ref{tab_wild} compares SDAIE$^\dagger$ against UnivFD and NPR in terms of detection accuracy. It is clear that SDAIE$^\dagger$ consistently attains the highest performance across all generators, which exhibits substantial visual and stylistic variability. NPR also maintains relatively strong accuracy on most sources, but trails SDAIE$^\dagger$ by $4$–$24$ percentage points, particularly on Qwen-Image and Tongyi. In contrast, UnivFD almost completely fails in this regime, underscoring the limitations of semantics-oriented CLIP features once emerging generators achieve high semantic plausibility and stylistic diversity.  Taken together with the benchmark results, these findings indicate that EXIF-induced, camera-intrinsic representations provide a robust and forward-compatible basis for detecting AI-generated images: they transfer reliably to unseen architectures, remain effective on real-world social media content, and avoid overfitting to artifacts specific to any particular generation family or deployment pipeline.

\parhead{Evaluation on Other Image Formats.} Because our detection pipeline is trained exclusively on photographic and AI-generated images, it is interesting to see its behavior on semantic formats that lie clearly outside this training distribution. Specifically, we collect $1,000$ hand-drawn or computer-aided cartoon images~\cite{zheng2020cartoon} and use them as an out-of-distribution test set. SDAIE$^\dagger$ identifies $98.7\%$ of these images as non-AI-generated, closely matching UnivFD and NPR ($99.5\%$ and $97.4\%$, respectively). The results therefore demonstrate that SDAIE$^\dagger$, like existing methods, behaves conservatively on such out-of-distribution content, assigning it to the non-AI-generated side of the binary decision rather than over-claiming AI attribution.

\begin{figure}[]
    \centering
    \includegraphics[width=\linewidth]{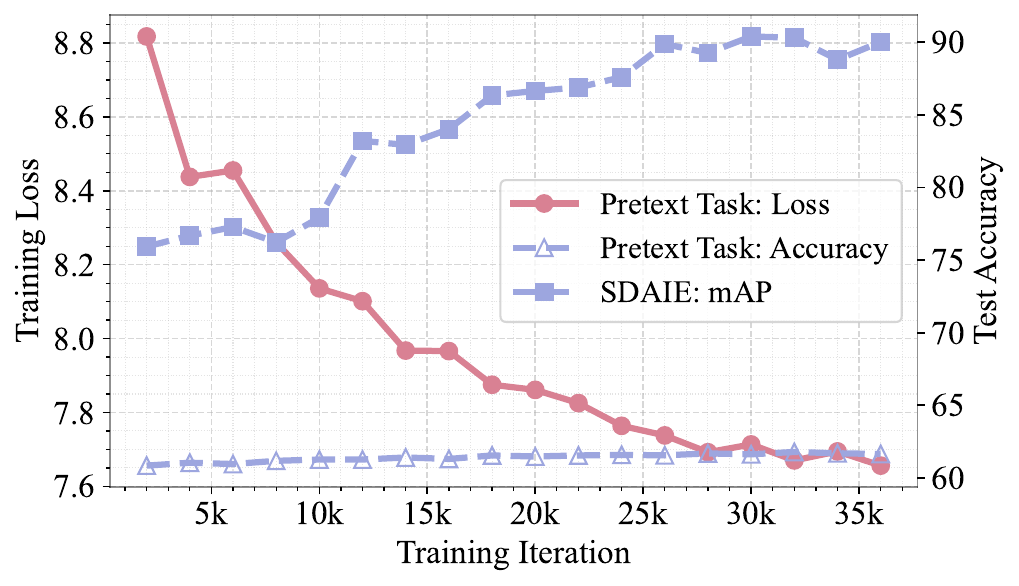}
    \caption{Effect of EXIF prediction accuracy on one-class detection performance of SDAIE.}
    \label{fig_ire_acc}

\end{figure}

\begin{figure}[!t]
    \centering
    \includegraphics[width=\linewidth]{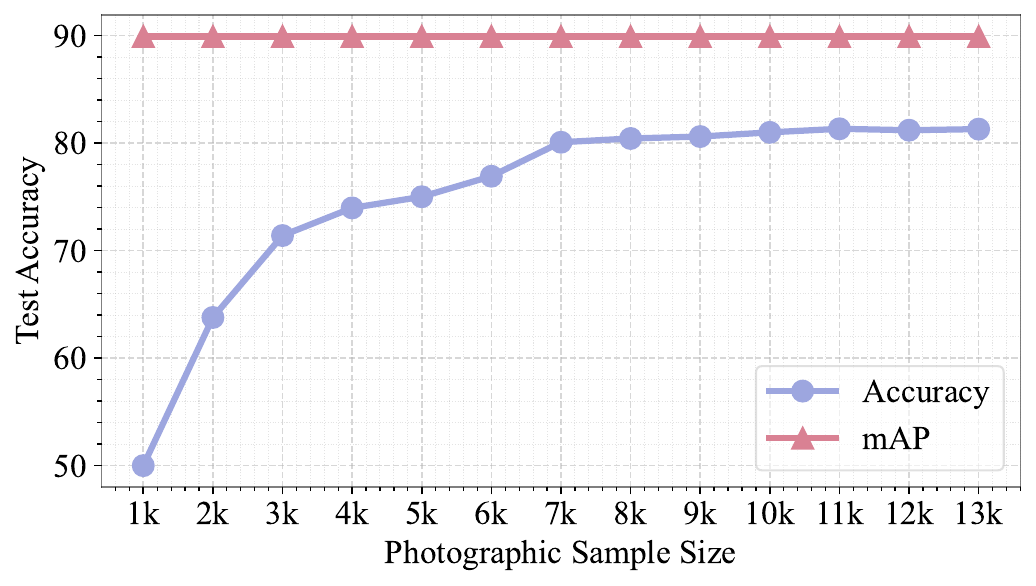}
    \caption{Effect of photographic training sample size on SDAIE.}
    \label{fig_ablation_real_image_number}
\end{figure}

\subsection{Ablation Studies}
We conduct a series of ablation studies to disentangle the contributions of the pretext task, network architecture, EXIF-induced regularization, and data augmentation to the overall performance of SDAIE and SDAIE$^\dagger$.

\parhead{Choice of Feature Extractor.} To demonstrate the importance of the EXIF-induced pretext task, we compare SDAIE with two alternative feature extractors trained on different pretext objectives: (1) the image encoder of CLIP, trained on large-scale image–text pairs, and (2) EAL~\cite{zheng2023exif}, trained on image–EXIF pairs. For all three extractors, one-class detection is implemented via a GMM in feature space.

Table~\ref{tab_ablation_pretext} shows that, averaged across all seventeen generators, SDAIE clearly outperforms both CLIP and EAL. CLIP’s performance deteriorates when AI-generated images become semantically plausible and diverse, reflecting its reliance on high-level semantics. EAL, while EXIF-aware, still lags behind SDAIE. We conjecture that learning from textual EXIF descriptions makes it difficult to capture fine-grained numerical relationships (\eg, between ISO 100 and ISO 400) embedded in long, information-dense text, whereas our direct classification and ranking over tag values offer a more precise way to encode camera-specific statistics.

\parhead{Impact of EXIF Prediction Accuracy on SDAIE.} We first examine how the quality of the EXIF-induced pretext task influences one-class detection. Fig.~\ref{fig_ire_acc} plots the evolution of the pretext training loss and test accuracy, together with the mAP of SDAIE, as a function of training iterations. As the cross-entropy loss of the pretext task decreases, its prediction accuracy improves, and the mAP of SDAIE also increases. Both curves gradually plateau once the number of iterations exceeds roughly $28,000$, showing that further optimizing the pretext task yields diminishing returns for downstream detection. This positive correlation confirms that better EXIF prediction directly translates into a more discriminative photographic feature space for one-class AI-generated image detection. 

\parhead{Effect of  Photographic Sample Size on SDAIE.}
SDAIE models the distribution of EXIF-induced features using a GMM fitted on a set of photographic images. We therefore study how its performance varies with the number of available photographic samples. Fig.~\ref{fig_ablation_real_image_number} reports the mean results across seventeen generators as we vary the number of training photographs from $1,000$ to $10,000$.

Even with only $1,000$ photographs, SDAIE attains an mAP of around $90\%$, indicating that the EXIF-induced feature extractor already creates a feature space where photographic and AI-generated images are largely separable. As the number of photographs increases, detection performance steadily improves; the average detection accuracy stabilizes around $80\%$ once more than $8,000$ images are used. Note that with very few training photographs, estimating a reliable likelihood threshold becomes challenging, and the resulting decision boundary can be unstable.

\parhead{Pretext Task Design.} We next assess the design choices in the EXIF-induced pretext task, focusing on how to handle continuous tags. We consider two alternatives: direct regression guided by the mean squared error versus the proposed pairwise ranking.
From our experiments, replacing regression with pairwise ranking yields consistent gains. For SDAIE, the accuracy improves from $79.55\%$ to $81.02\%$. For SDAIE$^\dagger$, the gains are even larger, with accuracy rising from $91.38\%$ to $94.30\%$. These results suggest that modeling relative ordering is more robust to the uneven quantization and noise typical of EXIF numerics than absolute regression, and better aligns the learned representations with camera-intrinsic trends.

\begin{table}[t]
  \centering
  \caption{Effect of network components and patch size on detection performance (Acc/mAP) of SDAIE and SDAIE$^\dagger$.}
  \label{tab_ablation_network}%
  \begin{tabular}{lcc}
    \toprule
    Configuration                 & SDAIE       & SDAIE$^\dagger$ \\
    \midrule
   w/o Patch Scrambling                     & 80.56/88.25 & 80.02/87.72     \\
    w/o High-Pass Filtering                    & 66.84/77.90 & 81.17/89.19     \\
    w/o Covariance Pooling           & 78.21/87.73 & 89.14/95.04     \\
    All (w/ Patch Size of $16\times16$) & 78.93/\textbf{92.63} & 84.09/90.16     \\
    \midrule
    All (w/ Patch Size of $64\times64$)                     & \textbf{81.02}/90.39 & \textbf{94.30}/\textbf{98.69} \\
    \bottomrule
  \end{tabular}
\end{table}

\begin{table}[]
\centering
\caption{Effect of high-pass filter design (see Fig.~\ref{fig_hpf}) on detection performance (Acc/mAP) of SDAIE and SDAIE$^\dagger$.}
\begin{tabular}{lccc}
\toprule
Filter & SDAIE & SDAIE$^\dagger$ \\
\midrule
Gabor Filters& 76.73/88.01 & 88.34/96.07 \\
Only (a) & 70.01/82.23 & 83.32/91.83 \\
(a) to (b) & 71.43/83.98 & 84.77/93.70 \\
(a) to (c) & 73.76/85.70 & 87.10/95.33 \\
(a) to (d) & 76.89/87.87 & 89.98/96.56 \\
(a) to (e) & 79.50/89.03 & 91.43/97.39 \\
(a) to (f) & 79.39/89.11 & 91.33/97.31 \\
\midrule
All ((a) to (g)) & \textbf{81.02}/\textbf{90.39} & \textbf{94.30}/\textbf{98.69} \\
\bottomrule
\end{tabular}
\label{tab_ablation_hpf_simplified}
\end{table}

\parhead{Network Architecture Design.} We then dissect the network architecture, which consists of scrambled patches, a bank of high-pass filters, a convolution encoder with covariance pooling, and a Transformer encoder. Table~\ref{tab_ablation_network} summarizes the effect of four key design choices: patch scrambling, high-pass filtering, covariance pooling, and patch size.

When all three modules are enabled with a patch size of $64\times 64$, SDAIE and SDAIE$^\dagger$ achieve the best performance. Reducing the patch size to $16\times 16$ significantly degrades SDAIE$^\dagger$, suggesting that very small patches fail to preserve sufficient spatial structure for reliably capturing camera-intrinsic residual patterns. Removing patch scrambling (by adding positional encoding~\cite{shaw2018self}), high-pass filtering, or covariance pooling (replaced with average pooling) each leads to noticeable drops in both accuracy and mAP, confirming that these components are complementary rather than redundant.

\parhead{Detailed Impact of High-Pass Filters.} Because high-pass filtering is important to our design, we further analyze its effect in isolation. Our high-pass filter bank comprises $30$ filters derived from seven prototype kernels (see Fig.~\ref{fig_hpf}). Table~\ref{tab_ablation_hpf_simplified} reports detection performance as we progressively add filter groups. Both SDAIE and SDAIE$^\dagger$ benefit monotonically from increasing the number of filters, with the full bank achieving the highest results. This trend suggests that a diverse set of structural and directional filters is essential for exposing subtle residual patterns left by camera pipelines versus synthesis processes.

We also compare our high-pass filtering scheme against Gabor filters~\cite{daugman1985uncertainty}. The latter yields inferior performance, confirming that the classical forensic filters of Fridrich and Kodovsk\'{y}~\cite{fridrich2012rich} are better suited to modeling camera-intrinsic traces in high-frequency residuals than generic orientation–frequency decompositions.

\begin{table*}[]
\centering
  \caption{Effect of data augmentation on the accuracy–robustness trade-off for NPR and SDAIE$^\dagger$ (Acc/mAP).}
\begin{tabular}{ccccccccc}
\toprule
\multicolumn{1}{c}{\multirow{2}{*}{Data Augmentation}} & \multicolumn{2}{c}{Clean}     & \multicolumn{2}{c}{JPEG Compression}      & \multicolumn{2}{c}{Gaussian Blurring} & \multicolumn{2}{c}{Downsampling} \\
\cmidrule{2-9}
\multicolumn{1}{c}{}                                    & NPR         & SDAIE$^\dagger$ & NPR         & SDAIE$^\dagger$ & NPR           & SDAIE$^\dagger$   & NPR           & SDAIE$^\dagger$  \\
\midrule
\xmark                                                & 92.14/98.21 & 76.48/88.71     & 50.09/49.06 & 49.97/49.72     & 52.96/53.31   & 60.10/69.89       & 76.91/80.62   & 59.83/62.19      \\
\checkmark                                            & 87.54/95.36 & 94.30/98.69     & 72.85/77.12 & 78.47/84.38     & 73.49/77.62   & 76.68/87.01       & 70.30/79.40   & 79.52/86.70      \\
\bottomrule
\end{tabular}
 \label{tab_ablation_aug}%
\end{table*}

\begin{figure}[!t]
    \centering
    \includegraphics[width=\linewidth]{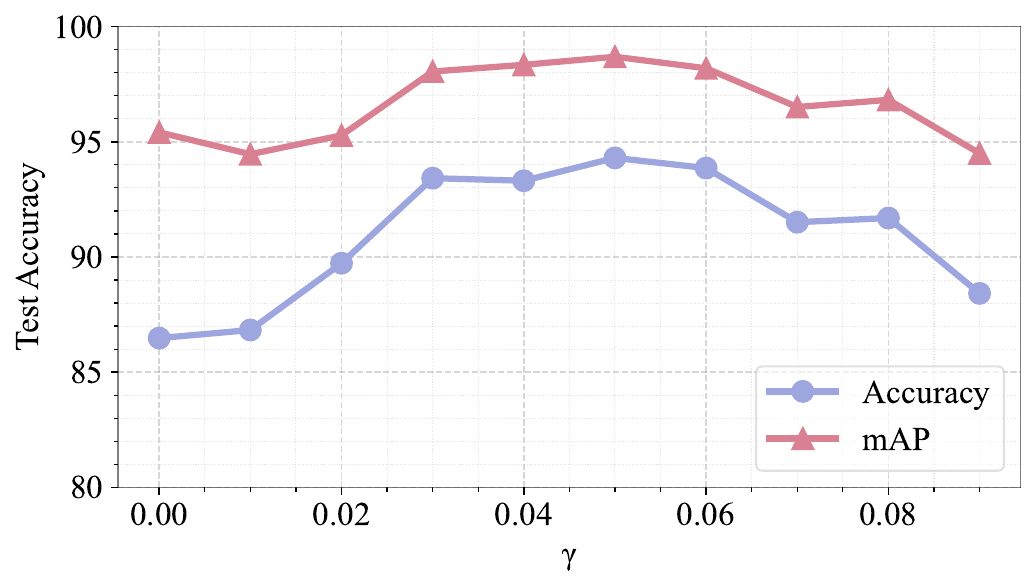}
    \caption{Effect of EXIF-induced regularization strength on SDAIE$^\dagger$.}
    \label{fig_ablation_alpha}
\end{figure}

\parhead{Effect of the EXIF-induced Regularizer on SDAIE$^\dagger$.} In the binary SDAIE$^\dagger$, the EXIF-induced feature extractor serves as a regularizer that encourages the detector to preserve camera-intrinsic representations while learning to discriminate between photographs and ProGAN-generated images. The strength of this guidance is governed by $\gamma$ in Eq.~\eqref{eq_binary_classification}.

Fig.~\ref{fig_ablation_alpha} shows detection performance for different $\gamma$. When $\gamma= 0$ (\ie, random initialization with no regularization), the accuracy decreases by roughly $8\%$ compared with the best setting, confirming that na\"{i}vely training a binary classifier on ProGAN-generated images versus photographs tends to overfit to generator-specific cues. As $\gamma$ increases, both accuracy and mAP improve, reaching their peak around $\gamma=0.05$. Further increasing $\gamma$ beyond $0.7$ constrains the classifier too heavily to the pretext representations, leaving insufficient flexibility to adapt to the binary detection task and causing performance to drop.

We also consider a fine-tuning baseline in which a classification head is appended directly to the self-supervised EXIF-induced feature extractor and the entire model is jointly optimized. This strategy achieves an accuracy of $91.78\%$, which remains notably below the best performance of SDAIE$^\dagger$ ($94.30\%$). These results emphasize that explicit representation alignment through regularization yields a better trade-off between preserving camera-intrinsic cues and adapting to the supervised detection objective than simple end-to-end fine-tuning.

\parhead{Effect of Data Augmentation.} We finally analyze how data augmentation affects the balance between accuracy and robustness for NPR and SDAIE$^\dagger$ under three benign perturbations: JPEG compression, Gaussian blurring, and downsampling (see Table~\ref{tab_ablation_aug}). For SDAIE$^\dagger$, augmentation yields consistent gains across both clean and perturbed settings. This signifies that EXIF-induced features can simultaneously benefit from augmentation in terms of generalization to unseen clean images and robustness to post-processed images, without incurring a noticeable trade-off.

For NPR, the effect of augmentation is more delicate and highlights an explicit accuracy–robustness trade-off. Without augmentation, NPR attains high accuracy and mAP on clean inputs, but is extremely fragile to JPEG compression and Gaussian blurring. Introducing the same augmentations substantially improves robustness to such perturbations, but these gains come at the cost of reduced performance on clean and downsampled images. In other words, for NPR, augmentation shifts the decision boundary toward invariance to post-processing, but simultaneously weakens its strong bias toward upsampling-induced local correlation patterns that drive its excellent clean-image performance.

\section{Conclusion and Discussion}
We have described a self-supervised feature extractor that leverages photographic EXIF metadata to guide the learning of camera-intrinsic representations. The pre-trained extractor maps spatial images into a compact feature space in which photographs and AI-generated images are well separated. Extensive experiments on both one-class and binary detection settings show that our EXIF-induced detectors achieve consistently superior generalization to unseen generators and improved robustness to common post-processing operations. 

Despite these promising results, several important questions remain open. First, the current pretext task is designed to predict EXIF-related properties and is therefore only implicitly aligned with the downstream detection objective. A more explicit, automated alignment between pretext and detection tasks~\cite{zou2025bi} may further improve the transferability of the learned representations. 

Second, aligning with previous studies~\cite{tan2024rethinking,tan2023learning}, our feature extractor and downstream detectors still exhibit a notable reliance on high-frequency traces, which can be fragile under common post-processing operations. Future work should therefore investigate architectures, regularization schemes, and data augmentation strategies that promote a more balanced use of multi-level, multi-scale cues, reducing dependence on high-frequency patterns while preserving sensitivity to subtle generative footprints. 

Third, we have considered a one-class detection setting in which any image deviating from the modeled photographic distribution is treated as anomalous (\ie, AI-generated). While this provides a simple mechanism to flag non-photographic content, it remains a coarse approximation of real-world usage. In practice, many non-photographic yet non-AI-generated images (\eg, graphics, paintings, or heavily edited content) are better handled as a distinct category rather than being lumped together with AI-generated imagery. Extending our framework to a more realistic out-of-distribution setting, with a third class explicitly modeling such non-photographic and non-AI-generated images, would better reflect real deployment conditions and help avoid overconfident predictions on genuinely out-of-domain inputs.

Finally, our evaluation inherits the common practice~\cite{tan2024rethinking,ojha2023towards,wang2023dire,tan2023learning,liu2022detecting,ju2022fusing} summarized in Table~\ref{tab_dataset}, where different generators are tested on heterogeneous photographic datasets. To enable more standardized and comparable benchmarking, an important community effort is to construct and adopt a large-scale, universal set of photographic images shared across generators and detection methods. Doing so would support more rigorous cross-generator comparisons and yield a clearer view of the strengths and limitations of competing detectors.

\bibliographystyle{IEEEtran}
\bibliography{main.bib}{}

\end{document}